\documentclass{ieeeaccess}
\usepackage[utf8]{inputenc}
\usepackage[english]{babel}
\usepackage{multirow}
\usepackage[dvipsnames]{xcolor}
\definecolor{accessblue}{RGB}{0,105,154}
\definecolor{greycolor}{cmyk}{0,0,0,1}
\usepackage{textcomp}
\usepackage{graphicx}
\graphicspath{{fig/}}
\usepackage{lipsum} 
\usepackage{amsmath,amsfonts,amssymb,amscd,bm}
\usepackage{algorithm}
\usepackage{algpseudocode}
\usepackage{csquotes}
\usepackage{booktabs}
\usepackage{tikz}
\usepackage{pgfplots}
\usepgfplotslibrary{groupplots,dateplot}
\usetikzlibrary{patterns,shapes,arrows,positioning,shapes.geometric,shapes.arrows}
\pgfplotsset{compat=newest}
\usepackage[normalem]{ulem}
\usepackage{subcaption}

\usepackage[style=ieee]{biblatex}
\usepackage{tabularx}
\usepackage{afterpage}
\usepackage{adjustbox}
\usepackage{hyperref}
\newcommand{\kpi}{y}
\usepackage{eso-pic}
\AddToShipoutPicture*{\tiny\sffamily\raisebox{0.6cm}{\hspace{1.3cm}\fbox{\parbox{\textwidth}{\copyright 2021 IEEE.  Personal use of this material is permitted.  Permission from IEEE must be obtained for all other uses, in any current or future media, including reprinting/republishing this material for advertising or promotional purposes, creating new collective works, for resale or redistribution to servers or lists, or reuse of any copyrighted component of this work in other works.}}}}

\doi{10.1109/ACCESS.2021.3053856}
\def\thevol{xx}
\def\theYear{2021}
\begin{document}
\title{Deep Learning-based Prediction of Key Performance Indicators for Electrical Machines}
\author{
\uppercase{Vivek Parekh}\authorrefmark{1}, 
\uppercase{Dominik Flore}\authorrefmark{2}, 
\uppercase{Sebastian Sch\"{o}ps}\authorrefmark{3}
}

\address[1]{Technische Universit\"{a}t Darmstadt, Computational Electromagnetics Group,\\ Schloßgartenstrasse 8, 64289 Darmstadt and Robert Bosch GmbH,\\Powertrain Solutions, Mechanical Engineering and Reliability,\\
70442 Stuttgart (e-mail: Vivek.Parekh@de.bosch.com)}
\address[2]{Robert Bosch GmbH, Powertrain Solutions, Mechanical Engineering and Reliability,\\
70442 Stuttgart, Germany (e-mail: Dominik.Flore@de.bosch.com)}
\address[3]{Technische Universit\"{a}t Darmstadt, Computational Electromagnetics Group,\\Schlossgartenstrasse 8, 64289 Darmstadt, Germany(e-mail: sebastian.schoeps@tu-darmstadt.de)}

\markboth
{V. Parekh, D. Flore, S. Sch\"{o}ps: Deep Learning-based Prediction of Key Performance Indicators for Electrical Machine}
{V. Parekh, D. Flore, S. Sch\"{o}ps: Deep Learning-based Prediction of Key Performance Indicators for Electrical Machine}

\corresp{Corresponding author: Vivek Parekh (e-mail: Vivek.Parekh@de.bosch.com).}

\begin{abstract} 
The design of an electrical machine can be quantified and evaluated by Key Performance Indicators (KPIs) such as maximum torque, critical field strength, costs of active parts, sound power, etc. Generally, cross-domain tool-chains are used to optimize all the KPIs from different domains (multi-objective optimization) by varying the given input parameters in the largest possible design space. This optimization process involves magneto-static finite element simulation to obtain these decisive KPIs. It makes the whole process a vehemently time-consuming computational task that counts on the availability of resources with the involvement of high computational cost. In this paper, a data-aided, deep learning-based meta-model is employed to predict the KPIs of an electrical machine quickly and with high accuracy to accelerate the full optimization process and reduce its computational costs. The focus is on analyzing various forms of input data that serve as a geometry representation of the machine. Namely, these are the cross-section image of the electrical machine that allows a very general description of the geometry relating to different topologies and the the classical way of scalar geometry parametrizations. The impact of the resolution of the image is studied in detail. The results show a high prediction accuracy and proof that deep learning-based meta-models are able to minimize the optimization time. The results also indicate that the prediction quality of an image-based approach can be made comparable to the classical way based on scalar parameters. 
\end{abstract}


\titlepgskip=-15pt
\maketitle

\section{Introduction}
\subsection{Motivation}

    An electrical machine is a paramount part of an electrical drive. The automotive industry currently favors permanent magnet synchronous machines (PMSM) due to their numerous advantages like greater power density, high efficiency, broad speed range, large torque-current ratio etc., see for example \cite{2006IJTIA.126..168K,4451077,6352905}. Usually, the use of materials such as neodymium-iron-boron magnets, copper, and electrical steel makes a significant contribution to the cost of the electrical machine. Only the smaller part of the final cost is due to the added value.  This is why numerical optimization of the active parts (rotor, magnets, stator, winding), i.e., minimizing material usage, can decrease the costs dramatically, before the PMSM is manufactured. Especially when the design space (e.g. winding topology, current/voltage range, geometry parameter range, material parameter) is large. For PMSMs, a large design space requires approximately 50-100 input parameters, the impact of which must be evaluated and optimized for around 10-20 Key Performance Indicators (KPIs). In such a large design space, multi-objective optimization is very time consuming and thus cost-intensive if the use of finite element (FE) models is needed to extract the KPIs. Meta-models (also known as surrogate models, kriging) are often fed to the optimizer in order to overcome this and to adapt to continuously reduced development cycles in the automotive industry. This enables optimizations in a large design space in short time.

    \begin{figure}
		\centering
        \begin{subfigure}[b]{.33\linewidth}
            \centering
            \scalebox{0.5}{%
		        \begin{tikzpicture}[y=0.80pt, x=0.80pt, yscale=-1, xscale=1, inner sep=0pt, outer sep=0pt]
\draw[fill=lightgray] (0.6600,16.7300) -- (58.0200,152.5700) .. controls (76.4100,144.3800) and (102.8100,144.3800) .. (118.4700,154.2100) -- (179.8300,16.9200) .. controls (122.1100,-4.7700) and (56.7400,-5.1200) .. (0.6600,16.7300) -- cycle;
\draw[fill=white] (22.4800,29.8900) .. controls (31.5700,34.2400) and (53.7500,41.4500) .. (70.3400,47.7700) .. controls (71.0500,48.4600) and (73.3500,51.9000) .. (78.5100,51.1300) .. controls (87.2700,49.8200) and (84.4200,33.1800) .. (80.0800,30.9600) -- (70.0600,30.4000) .. controls (70.0600,30.4000) and (38.0400,18.4200) .. (27.7300,14.7800) .. controls (15.3100,10.4000) and (19.2300,28.3300) .. (22.4800,29.8900) -- cycle;
\draw[fill=NavyBlue,cm={{0.9387,0.3449,-0.3449,0.9387,(14.0924,-15.5848)}},rounded corners=0.0000cm] (26.6000,26.0600) rectangle (75.1100,37.5800);
\draw[fill=white] (155.5200,30.6800) .. controls (146.4300,35.0300) and (124.2500,42.2400) .. (107.6600,48.5600) .. controls (106.9500,49.2500) and (104.6500,52.6900) .. (99.4900,51.9200) .. controls (90.7300,50.6100) and (93.5800,33.9700) .. (97.9200,31.7500) -- (107.9600,31.1900) .. controls (107.9600,31.1900) and (139.9800,19.2100) .. (150.2900,15.5700) .. controls (162.6900,11.1900) and (158.7700,29.1200) .. (155.5200,30.6800) -- cycle;
\draw[fill=NavyBlue,cm={{-0.9387,0.3449,-0.3449,-0.9387,(257.7451,19.3624)}},rounded corners=0.0000cm] (102.9000,26.8500) rectangle (151.4100,38.3700);
\end{tikzpicture}
		    }    
            \caption{Single V}\label{fig:RTOP-a}
        \end{subfigure}%
         \begin{subfigure}[b]{.33\linewidth}
            \centering
            \scalebox{0.5}{%
		        \begin{tikzpicture}[y=0.80pt, x=0.80pt, yscale=-1, xscale=1, inner sep=0pt, outer sep=0pt]
\draw[fill=lightgray] (0.6600,16.7300) -- (58.0200,152.5700) .. controls (76.4100,144.3800) and (102.8100,144.3800) .. (118.4700,154.2100) -- (179.8300,16.9200) .. controls (122.1100,-4.7700) and (56.7400,-5.1200) .. (0.6600,16.7300) -- cycle;
\draw[fill=white] (154.1800,53.3600) .. controls (145.0900,57.7100) and (122.9100,64.9200) .. (106.3200,71.2400) .. controls (105.6100,71.9300) and (105.3500,75.1200) .. (100.1500,74.6000) .. controls (92.8900,73.8800) and (92.2400,57.6500) .. (96.5800,55.4300) -- (103.4700,55.1400) .. controls (103.4700,55.1400) and (138.6400,41.8900) .. (148.9500,38.2500) .. controls (161.3500,33.8700) and (157.4300,51.8000) .. (154.1800,53.3600) -- cycle;
\draw[fill=NavyBlue,cm={{-0.9387,0.3449,-0.3449,-0.9387,(264.2023,63.2754)}},rounded corners=0.0000cm] (102.2200,49.3800) rectangle (150.7300,60.9000);
\draw[fill=white] (138.0600,26.2700) .. controls (128.9700,30.6200) and (121.4100,35.1100) .. (105.2000,43.1500) .. controls (104.4900,43.8400) and (104.9400,47.2200) .. (99.7100,47.3300) .. controls (94.0000,47.4400) and (92.1100,28.5600) .. (96.4500,26.3400) -- (101.5800,26.6700) .. controls (101.5800,26.6700) and (122.5100,14.8000) .. (132.8300,11.1600) .. controls (145.2300,6.7800) and (141.3100,24.7200) .. (138.0600,26.2700) -- cycle;
\draw[fill=NavyBlue,cm={{-0.8906,0.4548,-0.4548,-0.8906,(233.2893,-0.4777)}},rounded corners=0.0000cm] (101.6800,22.0600) rectangle (131.7300,33.5800);
\draw[fill=white] (26.3100,53.3600) .. controls (35.4000,57.7100) and (57.5800,64.9200) .. (74.1700,71.2400) .. controls (74.8800,71.9300) and (75.1400,75.1200) .. (80.3400,74.6000) .. controls (87.6000,73.8800) and (88.2500,57.6500) .. (83.9100,55.4300) -- (77.0200,55.1400) .. controls (77.0200,55.1400) and (41.8500,41.8900) .. (31.5400,38.2500) .. controls (19.1400,33.8700) and (23.0600,51.8000) .. (26.3100,53.3600) -- cycle;
\draw[fill=NavyBlue,cm={{0.9387,0.3449,-0.3449,0.9387,(22.3294,-15.2481)}},rounded corners=0.0000cm] (29.7700,49.3800) rectangle (78.2800,60.9000);
\draw[fill=white] (42.4400,26.2700) .. controls (51.5300,30.6200) and (59.0900,35.1100) .. (75.3000,43.1500) .. controls (76.0100,43.8400) and (75.5600,47.2200) .. (80.7900,47.3300) .. controls (86.5000,47.4400) and (88.3900,28.5600) .. (84.0500,26.3400) -- (78.9200,26.6700) .. controls (78.9200,26.6700) and (57.9900,14.8000) .. (47.6700,11.1600) .. controls (35.2700,6.7800) and (39.1900,24.7200) .. (42.4400,26.2700) -- cycle;
\draw[fill=NavyBlue,cm={{0.8906,0.4548,-0.4548,0.8906,(19.6344,-25.9704)}},rounded corners=0.0000cm] (48.7700,22.0600) rectangle (78.8200,33.5800);
\end{tikzpicture}
		    }    
            \caption{Double V}\label{fig:RTOP-b}
        \end{subfigure}%
        \begin{subfigure}[b]{.33\linewidth}
            \centering
            \scalebox{0.5}{%
		        \begin{tikzpicture}[y=0.80pt, x=0.80pt, yscale=-1, xscale=1, inner sep=0pt, outer sep=0pt]
\draw[fill=lightgray]  (0.6600,16.7300) -- (58.0200,152.5700) .. controls (76.4100,144.3800) and (102.8100,144.3800) .. (118.4700,154.2100) -- (179.8300,16.9200) .. controls (122.1100,-4.7700) and (56.7400,-5.1200) .. (0.6600,16.7300) -- cycle;
\draw[fill=white] (138.0000,25.3100) .. controls (128.9100,29.6600) and (118.0500,35.9200) .. (101.8300,43.9600) .. controls (101.1200,44.6500) and (101.4200,46.4200) .. (96.1900,46.5200) .. controls (90.4800,46.6300) and (91.7600,28.5000) .. (94.8900,27.1500) -- (102.9400,26.6700) .. controls (102.9400,26.6700) and (123.1400,16.2900) .. (132.3600,11.1700) .. controls (143.8600,4.7800) and (141.2500,23.7500) .. (138.0000,25.3100) -- cycle;
\draw[fill=NavyBlue,cm={{-0.8906,0.4548,-0.4548,-0.8906,(232.9036,-0.3849)}},rounded corners=0.0000cm] (101.4700,22.0600) rectangle (131.5200,33.5800);
\draw[fill=white] (164.2300,45.1600) .. controls (159.1000,47.6100) and (134.3100,65.5000) .. (134.3100,65.5000) .. controls (134.3100,65.5000) and (114.6900,71.2200) .. (107.4700,73.9800) .. controls (106.7600,74.6700) and (106.5200,80.4400) .. (101.3200,79.9200) .. controls (94.0600,79.2000) and (93.1900,64.5100) .. (97.5300,62.3000) -- (129.2900,53.4900) -- (158.9000,33.0000) .. controls (169.4500,27.1500) and (167.4800,43.6000) .. (164.2300,45.1600) -- cycle;
\draw[fill=NavyBlue] (136.5600,61.2700) -- (160.1300,45.4900) -- (155.3100,38.1100) -- (131.5400,53.8500) -- cycle;
\draw[fill=NavyBlue] (101.3100,73.2600) -- (128.3600,65.4700) -- (126.0800,56.4200) -- (99.0900,63.9300) -- cycle;
\draw[fill=white] (42.5000,25.3100) .. controls (51.5900,29.6600) and (62.4500,35.9200) .. (78.6700,43.9600) .. controls (79.3800,44.6500) and (79.0800,46.4200) .. (84.3100,46.5200) .. controls (90.0200,46.6300) and (88.7400,28.5000) .. (85.6100,27.1500) -- (77.5600,26.6700) .. controls (77.5600,26.6700) and (57.3600,16.2900) .. (48.1400,11.1700) .. controls (36.6300,4.7800) and (39.2500,23.7500) .. (42.5000,25.3100) -- cycle;
\draw[fill=NavyBlue,cm={{0.8906,0.4548,-0.4548,0.8906,(19.6568,-26.0631)}},rounded corners=0.0000cm] (48.9700,22.0600) rectangle (79.0200,33.5800);
\draw[fill=white] (16.2700,45.1600) .. controls (21.4000,47.6100) and (46.1900,65.5000) .. (46.1900,65.5000) .. controls (46.1900,65.5000) and (65.8100,71.2200) .. (73.0300,73.9800) .. controls (73.7400,74.6700) and (73.9800,80.4400) .. (79.1800,79.9200) .. controls (86.4400,79.2000) and (87.3100,64.5100) .. (82.9700,62.3000) -- (51.2100,53.4900) -- (21.6000,32.9900) .. controls (11.0400,27.1500) and (13.0200,43.6000) .. (16.2700,45.1600) -- cycle;
\draw[fill=NavyBlue] (43.9300,61.2700) -- (20.3700,45.4900) -- (25.1800,38.1100) -- (48.8200,53.7600) -- cycle;
\draw[fill=NavyBlue] (79.1800,73.2600) -- (52.1300,65.4700) -- (54.4100,56.4200) -- (81.4000,63.9300) -- cycle;
\end{tikzpicture}
		    }    
            \caption{VC-Design}\label{fig:RTOP-c}
        \end{subfigure}%
		\caption{Different rotor topologies for PMSM}
		\label{fig:RTOP}
   \end{figure}
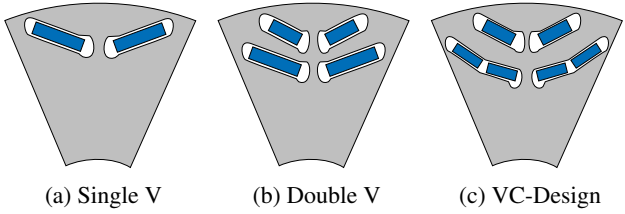%
    Another driver of this research work is to reuse simulation data or, in other words, to implement prior knowledge to a new optimization problem. Obviously, any optimization problem benefits if all the data (input parameters and output KPIs) of a previous problem were reused to direct the optimizer even faster into the optimal regions for the new problem. To date, transferring knowledge from one optimization to another is limited by the parametrization or the topology of an electrical machine. As shown in \autoref{fig:RTOP} different topologies of electric machines can be used to perform parametric optimization. However, the parametrization varies greatly for each topology and so does the input of the corresponding meta-model. Hence, a more general geometry description, such as an image can be used to transfer knowledge from one optimization to another.
    
    This paper investigates approaches to predict large numbers of KPIs of PMSM with a deep learning-based meta-model. It includes cross-domain KPIs, i.e. maximum torque, field strength, sound power level, cost of components, temperature level, etc. We address two questions. One is how accurately a classic scalar parameter-based meta-model can predict the KPIs with a given number of samples, and another is how the prediction accuracy of a more general and topology-invariant image-based meta-model performs (in dependency of the image resolution). 
    To answer these questions, two datasets are used which vary in design parameters and KPIs with different distribution of input data.

    The the paper is structured as follows: \autoref{sec:state} briefly discusses the current state of the art in industrial simulation and optimization workflows. \autoref{sec:model} details of datasets and KPIs. \autoref{sec:NAT} presents network architecture and training specifics. \autoref{sec:res} discuss the results and is followed by the conclusion.

\section{State of the art}\label{sec:state}

    Optimizing the design of an electrical machine involves cross-domain analysis, such as electromagnetic performance, stress and thermal behavior, etc., which essentially boils down to a multi-physics and multi-objective optimization (MOO) problem, e.g. \cite{6479303,Rosu_2017aa,8280552}. To carry out MOO, various optimization approaches are outlined for example in \cite{Ehrgott_2005aa,Di-Barba_2010aa,lei2017review,8972257,9119792}.

    \subsection{Multi-objective optimization}
\begin{figure}
		\centering
        \tikzstyle{block} = [rectangle, draw, fill=black!10, 
            text width=12em, text centered, rounded corners,
            minimum height=3em]
        \tikzstyle{line} = [draw, -latex']
        \begin{tikzpicture}[node distance=4.5em]
            \tikzset{font=\footnotesize}
            \node [block] (init) {Input parameters $\mathbf{p}$ with constraints (geometry, electrical and material)};
            \node [block, below of=init] (lhs) {Initial population by LHS $\mathbf{P}^{(0)}$};
            \node [block, below of=lhs] (maxwell) {Compute KPIs via \eqref{eq:kpi} from \eqref{eq:reducedmaxwell} for all $\mathbf{p}$$\in$$\mathbf{P}^{(i)}$};
            \node [block, below of=maxwell] (pareto) {Design evaluation and pareto front created};
            \node [block, below of=pareto] (swarm) {\textbf{Optimizer}: Creation of new designs $\mathbf{P}^{(i+1)}$ by selection, recombination and mutation};
            \node [left of=pareto, xshift=-4em] (repeat) {$i:=i+1$};
            \path [line] (init) -- (lhs);
            \path [line] (lhs) -- node[anchor=west] {$i=0$}(maxwell);
            \path [line] (maxwell) -- (pareto);
            \path [line] (pareto) -- (swarm);
            \path [line] (swarm) -| (repeat);
            \path [line] (repeat) |- (maxwell);
        \end{tikzpicture}
		\caption{Flowchart of the process for calculating KPIs}
		\label{fig:FCP}
   \end{figure}
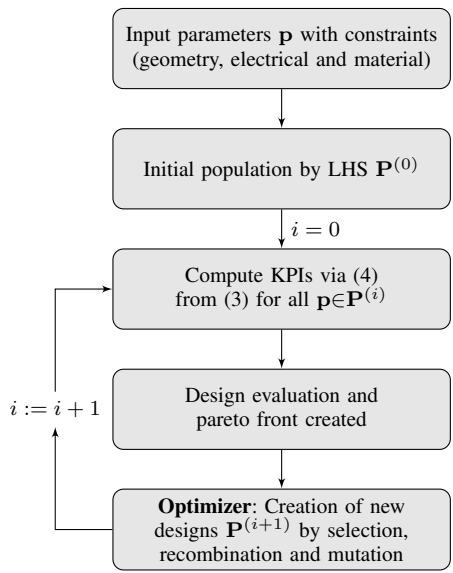

    The design of an electrical machine shall be obtained by MOO; its goal functions are the (possibly conflicting) KPIs $\kpi_j(\mathbf{p})\in\mathbb{R}$ with $j=1,\ldots,n_\mathrm{\kpi}$, e.g. maximum torque, maximum power, the tonality of a machine, torque ripple behavior, mass of the rotor, which depend on a parameter vector $\mathbf{p}\in\mathbb{R}^{n_\mathrm{p}}$ (geometry, material and electrical excitation). This optimization problem can be abstractly written as
    \begin{align}
        \label{eq:opt1}
        \min_\mathbf{p}\quad & \kpi_j(\mathbf{p}),        & j=1,\dots,n_\mathrm{K}\\
        \label{eq:opt2}
        \text{s.t.}\quad     & c_k(\mathbf{p}) \leq 0, & k=1,\dots,n_\mathrm{c}
    \end{align}
    where $c_k(\mathbf{p})$ may denote additional constraints, e.g. to avoid intersections of the geometry. 
    The simulation and optimization chain is illustrated in \autoref{fig:FCP}. 
    
    Most common multi\-objective optimizers solve (\ref{eq:opt1}-\ref{eq:opt2}) by creating first an initial population $\textbf{P}^{(0)}=\{\textbf{p}_1^{(0)},\ldots,\textbf{p}_{n_\textrm{LHS}}^{(0)}\}$, e.g., using Latin hypercube sampling (LHS) \cite{mckay2000comparison}. For each realization $\mathbf{p}\in\textbf{P}^{(0)}$ a finite element approximation of a magneto-static problem, e.g. on a 2D parameterized geometry $\Omega(\textbf{p})\subset\mathbb{R}^2$
    \begin{align}
        -\nabla\cdot\left(\nu\nabla A_z(\textbf{p})\right) = J_{\mathrm{src},z}(\textbf{p}) + \nabla \times \mathbf{M}(\textbf{p}) \cdot  \mathbf{e}_z, \label{eq:reducedmaxwell}
    \end{align}
    is required, \cite{Salon_1995aa,Merkel_2019aa}. Here, the (nonlinear) reluctivity is denoted by $\nu$, the z-component of the magnetic vector potential by $\mathbf{A}_z(\textbf{p})$, the current density by $J_{\mathrm{src},z}(\textbf{p})=\sum_k\chi_k i_k(\textbf{p})$ in terms of winding functions $\chi_k$ and currents $i_k(\textbf{p})$ \cite{Schops_2013aa}. The magnetization of the permanent magnets is taken into account by $\mathbf{M}(\textbf{p})$ and homogeneous Dirichlet boundary conditions $A_z=0$ are set on $\partial\Omega$. The computed electromagnetic fields are post-processed with various cross-domain tools and the end-results of this analysis are the KPIs used for optimization. They will be addressed by 
    \begin{align}
        \label{eq:kpi}
        \kpi_j^{(i)} := \kpi_j(A_z(\textbf{p}^{(i)}),\textbf{p}^{(i)}).
    \end{align}
    The pareto front consisting of the current optimal designs is created from those evaluations. Then, the optimizer generates new ensembles $\textbf{P}^{(i)}$ ($i>0$) e.g. with an evolutionary algorithm by selecting, recombining, and mutating \cite{Deb_2001aa}. The process repeats until convergence, see \autoref{fig:FCP}. The entire operation will take days or weeks depending on the availability of high performance computing resources. Meta-models which approximate the KPIs $\kpi_j$ by inexpensive surrogates $\tilde{\kpi}_j$ can overcome the computational burden by reducing the amount of evaluations of the costly FE problem \eqref{eq:reducedmaxwell}.
    
\subsection{Meta-model}
    Meta-modelling imitates the behavior of computationally expensive simulation model to evaluate desired objective functions as close as possible to the actual with being computationally cheap. Polynomial interpolation (also known as spectral method or polynomial chaos) is often used, in particular in the context of uncertainty quantification \cite{Xiu_2010aa,Clenet_2013aa,Bontinck_2016aa}. On the other hand, Kriging is another common approach to obtain a meta-model under suitable assumptions on priors \cite{forrester2008engineering}. In order to solve the optimization problem of electromagnetics in an inexpensive way, the possibility of meta-modelling with Kriging is for example proposed in \cite{4563368}. To optimize electromagnetic design by putting focus on achieving balance between exploitation and exploration during global optimum search with Kriging is proposed in \cite{article}. It is demonstrated in \cite{6556123}, how Kriging can be combined with evolutionary algorithms for MOO of PMSM design to lower the time-consuming computations. In \cite{7299302}, to enhance sensorless control capability along with torque behavior of the multi-objective surface mounted PMSM design for decreased optimization calculation time, a Kriging supported evolutionary algorithm is proposed.
                
    \begin{figure*}
     	\centering
     	\tikzset{every picture/.style={line width=0.75pt}} 

\begin{tikzpicture}[x=0.5pt,y=0.5pt]

\draw[fill=black,fill opacity=0.3] (0,0) -- (100,0) -- (100,100) -- (0,100) -- cycle;
\draw[] (-40,0) -- (140,0) -- (140,140) -- (-40,140) -- cycle;
\draw[fill=black,fill opacity=0.3] (250+0,0) -- (250+100,0) -- (250+100,100) -- (250+0,100) -- cycle;
\draw[] (250-40,0) -- (250+140,0) -- (250+140,140) -- (250-40,140) -- cycle;
\draw [<->] (-40,50) -- (  0,50) node [midway,anchor=south,font=\footnotesize] {$a$};

\draw [<->] (100,50) -- (140,50) node [midway,anchor=south,font=\footnotesize] {$a$};
\draw [<->] (50,100) -- (50,140) node [midway,anchor=east,font=\footnotesize] {$b$};
\draw [<->] (250-10,0) -- (250-10,100) node [midway,anchor=east,font=\footnotesize] {$d$};
\draw [<->] (250+0,100+10) -- (250+100,100+10)  node [midway,anchor=south,font=\footnotesize] {$e$};
\draw [<->] (-40,-10) -- (140,-10) node [midway,anchor=north,font=\footnotesize] {$w$};
\draw [<->] (250-40,-10) -- (250+140,-10) node [midway,anchor=north,font=\footnotesize] {$w$};
\draw [<->] (250+140+10,0) -- (250+140+10,140) node [midway,anchor=west,font=\footnotesize] {$h$};
\draw [<->] (140+10,0) -- (140+10,140) node [midway,anchor=west,font=\footnotesize] {$h$};
\end{tikzpicture}
        \captionsetup{justification=centering}
        \caption{Differently parameterized plates with the same image space ($e=w-2a$ and $d=h-b$)}
        \label{fig:plate}
    \end{figure*}
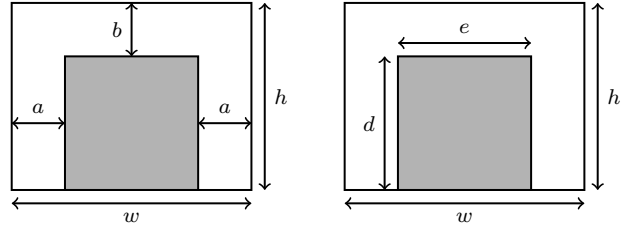
    
    At present, fast-paced developments in the domain of machine learning (ML), especially with deep learning, have unfolded new insight for complex non-linear function approximation for multi-output regression \cite{HORNIK1989359,liang2016why}. The convolutional neural network (CNN) which is a special class of artificial neural network has been used for decades now. The advantage of CNN is the extraction of features to represent the hierarchical expression of image form data \cite{726791} \cite{lecun2015deep}. The applications of CNN in various fields increased after the success in the image net challenge (classification and object detection via supervised learning) \cite{krizhevsky2012imagenet}. The ability to transfer pre-trained network knowledge from one application to another makes it prominent for faster training. It has shown good results even when training data is limited \cite{7426826}. 

    In the domain of electrical machines, deep learning applications are still in an early stage. The prediction of efficiency, speed and torque using DL based multi-regression was demonstrated for the performance analysis of PMSMs in \cite{8056321}. In \cite{gletter2019novel} it is shown how hybrid electric vehicles at system level can be optimized by modeling non-linear system behavior using neural networks. In another article, estimation of the magnetic field solution for different EM devices such as a coil in air, a transformer, and an interior permanent magnet (IPM) machine have been investigated by using deep CNNs \cite{8661767}. There is an application with recurrent neural network (RNN) and CNN for real time monitoring of high-fluctuating temperature inside PMSMs probed in \cite{8785109}. Deep neural networks were shown to work as torque predictors for different states of (steady or transient) interior PMSM drives \cite{8834195}. DL based on CNN has been shown to be effective for quick evaluation of electric motor performance in order to reduce FE analysis for the topology optimization \cite{sasaki2019topology}. This idea has been expanded for multi-objective topology optimization using Deep convolutional neural network (DCNN) \cite{8666164}. The efficiency map for a motor drive was computed with DL in \cite{8961095}. One of the recent work demonstrates a multi-layer perceptron as a meta-model for shape optimization of PMSMs \cite{You2020MOO}. In the recent past, a combination of the CNN based model and a reduced FE model were analyzed for accelerated optimizations in electromagnetics \cite{8957004}.

    \subsection{Comparison of parameter and image based meta-models}  
    The image-based approach is a very general way to access the performance of the electrical machine. The DL based meta-model becomes deterministic, once it is trained. The image-based DCNN model only considers the final image space in which it was trained. It is not concerned with how it is generated. So if we re-parameterize such that image space remains invariant, then it is possible to predict the KPIs by the same image based trained meta-model. However, this does not hold for (scalar) parameter based meta-models as any re-parametrization alters the input space. For example, as shown in \autoref{fig:plate}, two plates are differently parameterized but have the same image space. The plate 1 is generated with input scalar parameters $a, b$ while plate~2 is produced with parameters $d,e$ which are different. Now, if we train a parameter based meta-model with the input space of parameters $a, b$ and test with input space of parameters $d,e$ to make predictions, e.g. about the stiffness of plate, then it will not give correct prediction (if we do not find a suitable transformation). On the contrary, an image based trained DCNN model will still give correct predictions.

\section{Dataset generation}\label{sec:model} %

In this study, an optimization workflow is applied to generate a large amount of training data. Two datasets are considered. Each set consists of machine realizations, 
i.e., the parameter values $\mathbf{p}^{(i)}$, the Computer-Aided Design (CAD) models, which are then used for simulation and the corresponding KPIs $\kpi_j^{(i)}$. Details on two datasets can be found in the following subsections. Only a half pole and full pole cross-section are considered in both datasets since geometrical symmetry of the electrical machines can be exploited.
\subsection{Dataset~1}
\begin{table}
    \centering
    \captionsetup{justification=centering}
    \caption{Stator parameter detail dataset~1}
    \renewcommand{\arraystretch}{1.2}
    \footnotesize
    \begin{tabular}{@{}|l|l|r|r|c|@{}}
    \hline
             & \textbf{Parameter}       & \textbf{Min.}  & \textbf{Max.} & \textbf{Unit} \\ \hline
    $p_{ 1}$ & Tooth head overhang 1    & $  0.76$       & $  1.19$      & mm \\ \hline
    $p_{ 2}$ & Height of tooth head     & $ 12.41$       & $ 18.91$      & mm \\ \hline
    $p_{ 3}$ & Tangential groove width  & $  4.23$       & $  6.37$      & mm \\ \hline
    $p_{ 4}$ & Stator inner diameter    & $143.41$       & $158.34$      & mm \\ \hline
    $p_{ 5}$ & Tooth head overhang 2    & $  1.20$       & $  1.64$      & mm \\ \hline
    $p_{ 6}$ & Tooth width near air gap & $  5.05$       & $  8.60$      & mm \\ \hline
    $p_{ 7}$ & Iron length              & $160.49$       & $168.00$      & mm \\ \hline
    \end{tabular}%
    \label{tab:STPDS1}
\end{table}

\begin{table}
    \centering
    \captionsetup{justification=centering}
    \caption{Constant parameters}
    \renewcommand{\arraystretch}{1.2}
    \footnotesize
    \begin{tabular}{@{}|l|c|c|c|@{}}
    \hline
    \textbf{Parameter}       & \textbf{Dataset~1} & \textbf{Dataset~2} & \textbf{Unit} \\ \hline
    No pf pole pair          & $4$                & $4$                & -\\ \hline
    Stator type              & Asymmetric         & Asymmetric         & -\\ \hline
    Rotor  type              & VC-Design          & VC-Design          & -\\ \hline
    No of slots (stator)     & $48$               & $48$               & -\\ \hline
    Max. phase voltage       & $640$              & $640$              & V\\ \hline
    Max. phase current       & $480$              & $600$              & A\\ \hline
    Slots per pole per phase & $2$                & $2$                & -\\ \hline
    \end{tabular}%
    \label{tab:CP}
\end{table}

\begin{figure}
	\centering
	    \begin{subfigure}[b]{0.48\columnwidth}
	        \includegraphics[width=\columnwidth]{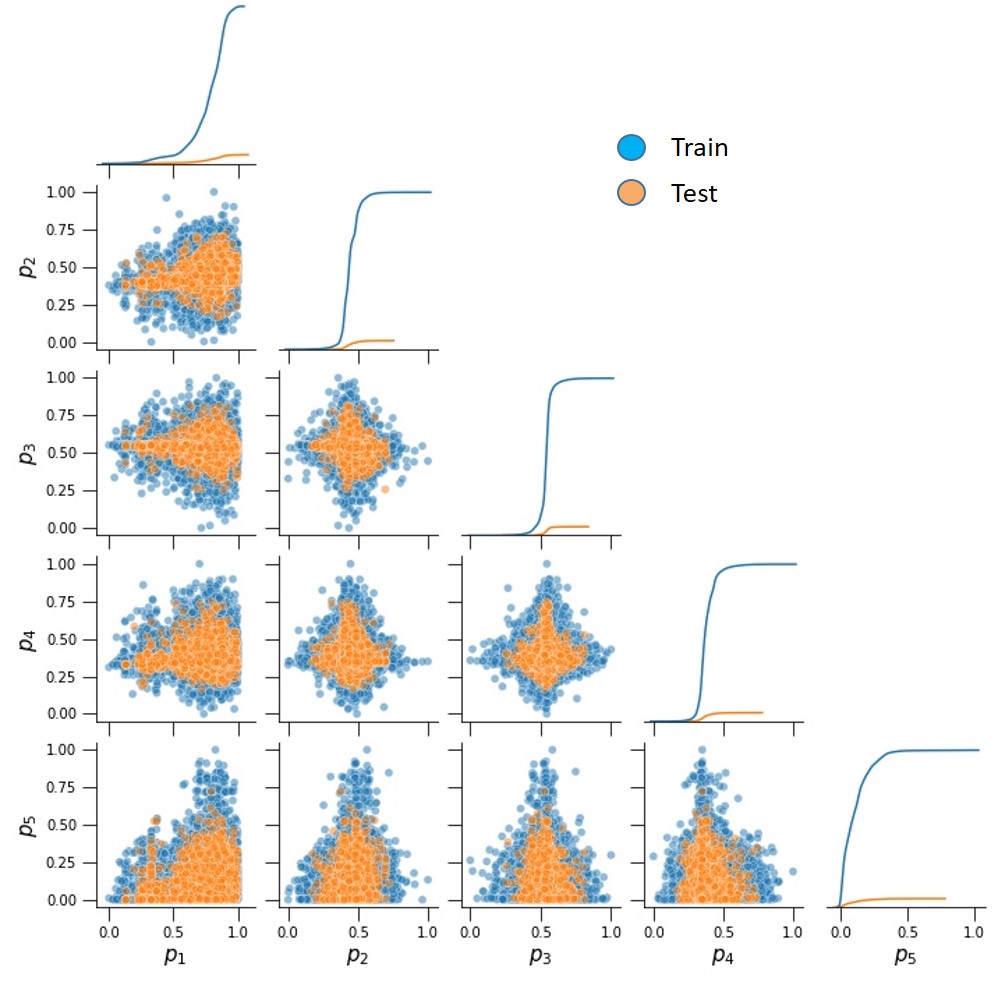}
	        \captionsetup{justification=centering}
	        \caption{Parameter distribution dataset~1}
            \label{fig:PDS1}
	    \end{subfigure}%
	    \begin{subfigure}[b]{0.48\columnwidth}
	        \includegraphics[width=\columnwidth]{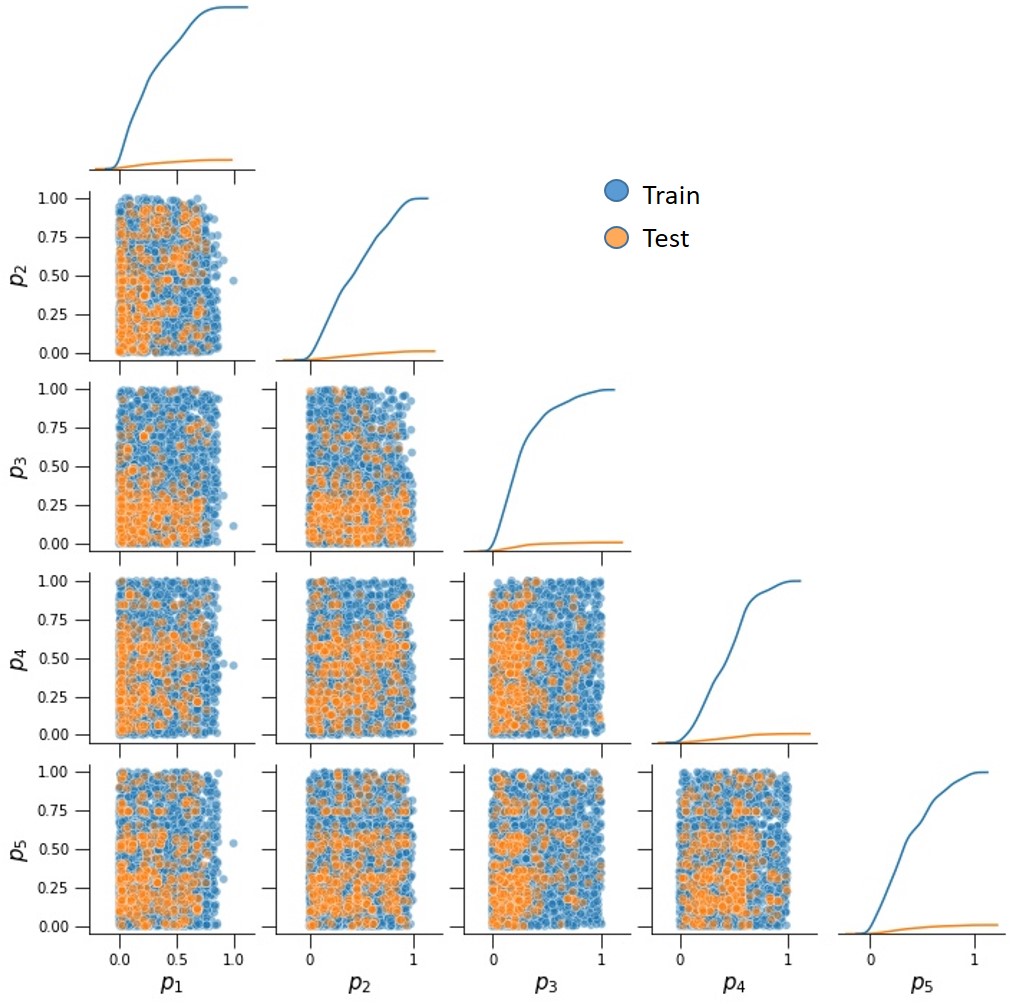}
	        \captionsetup{justification=centering}
	        \caption{Parameter distribution dataset~2}
            \label{fig:PDS2}
	    \end{subfigure}%
	    \\
        \begin{subfigure}[b]{0.48\columnwidth}
            \includegraphics[width=\columnwidth]{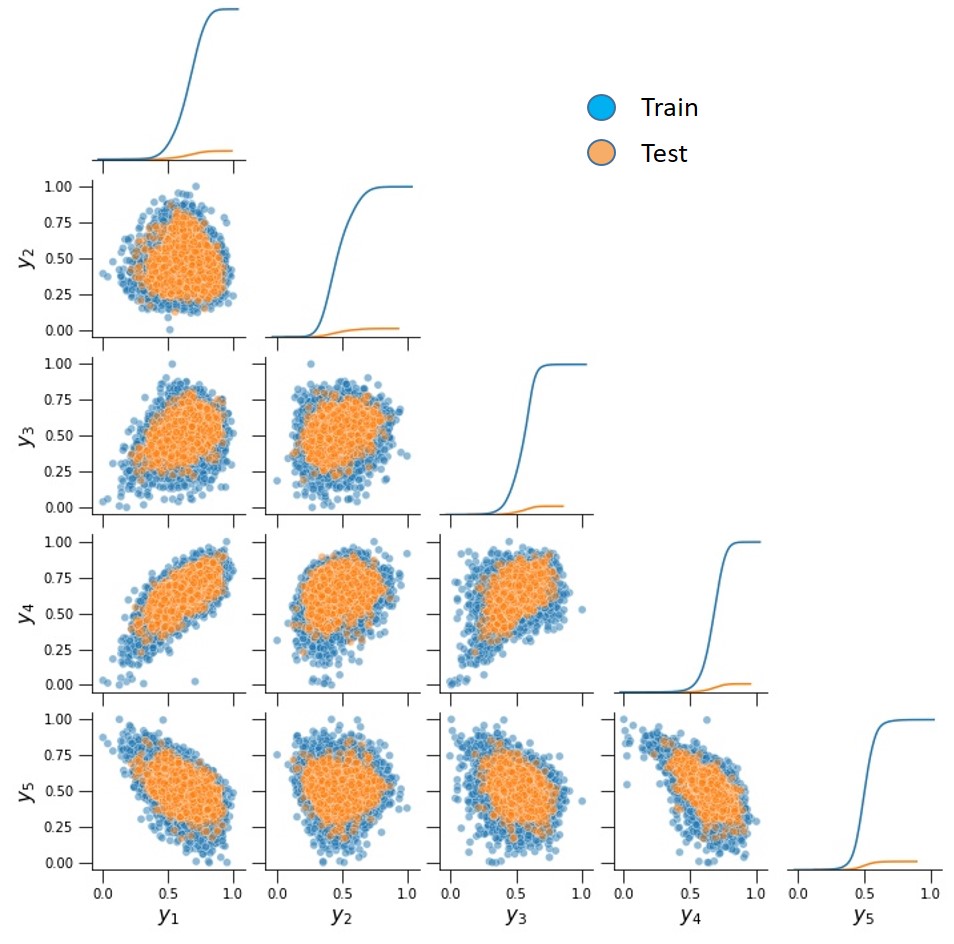}
            \captionsetup{justification=centering}
            \caption{KPI distribution datset 1}
            \label{fig:KPIDS1}
        \end{subfigure}%
        \begin{subfigure}[b]{0.48\columnwidth}
            \includegraphics[width=\columnwidth]{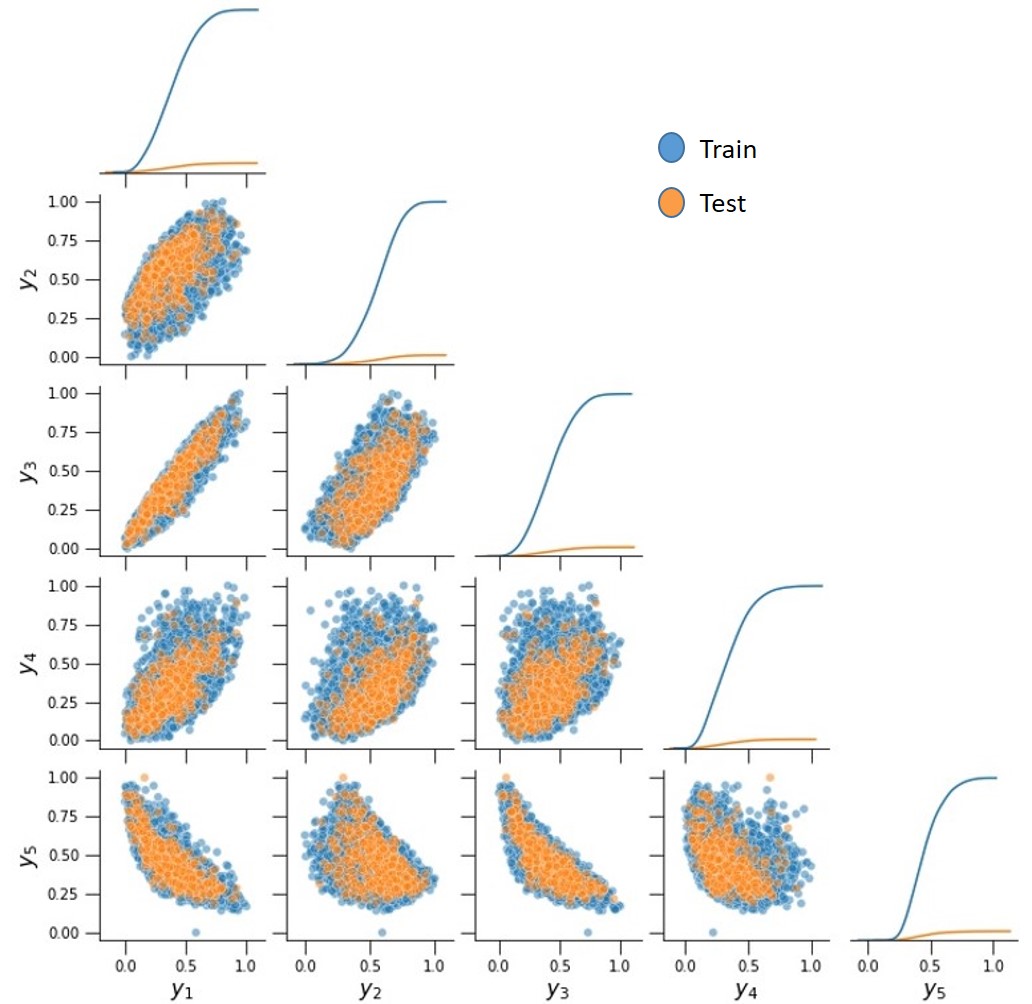}
            \captionsetup{justification=centering}
            \caption{KPI distribution dataset~2}
            \label{fig:KPIDS2}
        \end{subfigure}%
    \captionsetup{justification=centering}
    \caption{Parameter and KPI distributions}
    \label{fig:KPIDS}
\end{figure}

The rotor model takes into account $n_\mathrm{p}=49$ scalar parameters for the generation of samples. Seven important stator parameters that represent stator geometry information are detailed in \autoref{tab:STPDS1}. Other electrical parameters such as the number of slots, phase current, phase voltage, which remain constant during the data generation are given in \autoref{tab:CP}. Likewise, material features such as copper filling factor, remanence, and type of magnet cluster also remain invariant. The \autoref{tab:KPIDS1} gives a short description of the KPIs. The distributions in the spaces of parameters and KPIs of the model and its simulation results are visualized on affine 2D subspaces in \autoref{fig:PDS1} and \autoref{fig:KPIDS1}, respectively. The total number of samples produced using the \autoref{fig:FCP} process is $n_1=68099$ and the distribution of the input parameters is rather inhomogeneously distributed. 

The pre-processing of the data involves the transformation of the parametrized CAD model into a rectangular pixelized image in which each pixel has a unique identifier value related to the electrical machine component (air: 0, metal: 1, magnet: 2). The resulting cross-sectional images of a half pole of the rotors, shown for one example in \autoref{fig:DS1}, are used for the images-based training. 

\begin{table}
    \centering
    \captionsetup{justification=centering}
    \caption{KPIs information dataset~1}
    \footnotesize
    \renewcommand{\arraystretch}{1.2}
    \begin{tabular}{@{}|r|l|c|@{}}
    \hline
                & \textbf{KPI}                        & \textbf{Unit} \\ \hline
    $\kpi_{ 1}$ & Costs   of  active parts            & Euro          \\ \hline                 
    $\kpi_{ 2}$ & Critical   field strength,          & kA/m          \\ \hline                                                 $\kpi_{ 3}$ & Maximum   torque of machine         & Nm            \\ \hline  
    $\kpi_{ 4}$ & Maximum   power of machine          & W             \\ \hline                 
    $\kpi_{ 5}$ & Weighted   efficiency value         & \%            \\ \hline            
    $\kpi_{ 6}$ & Maximum torque-ripple               & Nmp           \\ \hline                
    $\kpi_{ 7}$ & Torque-ripple behavior of machine   & -             \\ \hline              
    $\kpi_{ 8}$ & Inverter   losses                   & W             \\ \hline
    $\kpi_{ 9}$ & Sound   power level of machine      & dBA           \\ \hline              
    $\kpi_{10}$ & Maximum   magnet temperature        & K             \\ \hline                  
    $\kpi_{11}$ & Maximum   winding temperature       & K             \\ \hline                  
    \end{tabular}%
    \label{tab:KPIDS1}
\end{table}

\begin{figure*}
		\centering
		\input{fig/dataset_1_vis}
		\vspace{-1em}
		\captionsetup{justification=centering}
    	\setlength{\abovecaptionskip}{15pt plus 3pt minus 2pt}
		\caption{Dataset~1 visualization}
		\label{fig:DS1}
\end{figure*}%
          
\begin{table*}
    \centering
    \captionsetup{justification=centering}
    \caption{Pixel Resolution Detail concerning to geometry parameter variation with dataset~1}
    \footnotesize
    \renewcommand{\arraystretch}{1.2}
    \begin{tabular}{@{}|l|r|r|r|r|r|r|r|r|@{}}
    \hline
    \multirow{3}{*}{~} &
      \multirow{3}{*}{\begin{tabular}[c]{@{}c@{}}Min\\    {[}mm{]}\end{tabular}} &
      \multirow{3}{*}{\begin{tabular}[c]{@{}c@{}}Max\\    {[}mm{]}\end{tabular}} &
      \multicolumn{6}{c|}{Image resolution in pixels,  X-direction=50mm, Y-direction=79mm} \\
       &            &            & \multicolumn{3}{c|}{Precision [mm/pixel]} & \multicolumn{3}{c|}{Pixel value}\\ 
       &            &            & $136\times216$ & $272\times432$ & $544\times864$ & $136\times216$ & $272\times432$ & $544\times864$ \\ \hline
    $p_1$ & $  0.8528$ & $  1.4895$ & $0.3676$       & $0.1838$       & $0.0919$       & $ 2$           & $ 4$           & $  7$          \\ \hline
    $p_2$ & $  7.1938$ & $  9.4859$ & $0.3676$       & $0.1838$       & $0.0919$       & $ 7$           & $ 13$           & $ 26$          \\ \hline
    $p_3$ & $  6.6480$ & $ 12.967$ & $0.3676$       & $0.1838$       & $0.0919$       & $18$           & $35$           & $ 70$          \\ \hline
    $p_4$ & $141.6990$ & $155.3637$ & $0.3676$       & $0.1838$       & $0.0919$       & $38$           & $75$           & $149$          \\ \hline
    \end{tabular}%
    \label{tab:IRDS1}
\end{table*}

\bigskip

One interesting observation is that the parameter-based KPI estimation is only dependents on the scalar parameters such that a even a tiny change of a single parameter will in general lead to a different prediction while the image-based model relies on the image accuracy.  In an initial examination, a resolution of $136\times216$\,pixels for the geometrical domain $79$mm$\times50$\,mm is selected. This results in $0.36$\,mm/pixel which is larger than the minimum  variation of any input scalar parameter, in other words, it takes approx. $3$\,pixels to indicate a change of $1$mm in a geometry parameter. This obviously affects the sensitivity of the network. Thus, if the pixel precision is increased, the interpretation of variations in the geometry parameters is enhanced. Four rotor parameters, which vary from minimum to maximum range, are set out in \autoref{tab:IRDS1}. The last three columns in the table give data about how many number of pixels required to show variation in the geometry for a unit length. Eventually, we compare three resolution values, $136\times216$, $272\times432$ and $544\times864$\,pixels.

\subsection{Dataset~2}

\begin{figure*}
		\centering
		\input{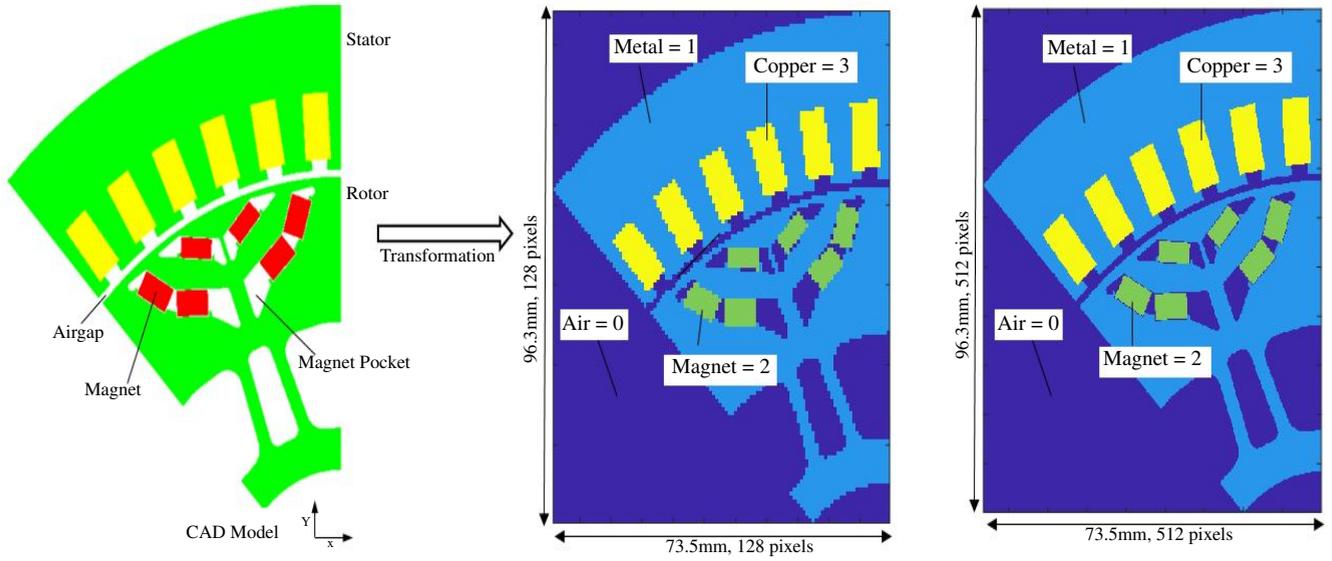}
		\vspace{-1em}
		\captionsetup{justification=centering}
    	\setlength{\abovecaptionskip}{15pt plus 3pt minus 2pt}
		\caption{Dataset~2 visualization}
		\label{fig:DS2}
\end{figure*}

\autoref{fig:DS2} illustrates dataset~2. It differs from dataset~1 by its parametrization, which is now given in terms of $n_\mathrm{p}=12$ values, and consequently by the image form. In the geometry image the stator and the rotor full pole cross-sections are visible. The transformed pixel-matrix includes an additional identifier tag value 3 to show copper material. Twelve major variable scalar parameters (rotor and stator) with their respective ranges are specified in \autoref{tab:PDS2}. Other constant parameters similar to dataset~1 appear in \autoref{tab:CP}. \autoref{tab:KPIDS2} details the respective KPIs. In the \autoref{fig:KPIDS2} and \autoref{fig:PDS2} respectively, the joint distribution of KPIs and scalar parameters is shown. The distribution of input parameters is almost uniform. For this dataset, the total number of samples generated is $n_2=7744$ 

 \begin{figure}
	\centering
    \begin{tikzpicture}
 
    \definecolor{color0}{rgb}{0.12156862745098,0.466666666666667,0.705882352941177}
    \definecolor{color1}{rgb}{1,0.498039215686275,0.0549019607843137}
    \definecolor{color2}{rgb}{0.172549019607843,0.627450980392157,0.172549019607843}
    \definecolor{color3}{rgb}{0.83921568627451,0.152941176470588,0.156862745098039}

    \begin{axis}[legend cell align={left},
                width=0.9\linewidth,
                legend style={fill opacity=0.8, draw opacity=1, text opacity=1, draw=white!80!black},
                tick align=outside,
                tick pos=left,
                x grid style={white!69.0196078431373!black},
                xlabel={Epochs},
                xmin=-2.5, xmax=38.5,
                xtick style={color=black},
                y grid style={white!69.0196078431373!black},
                ylabel={Mean squared error},
                ymin=0.0029523, ymax=0.1315637,  
                yticklabels={},               
                extra y ticks={0.02,0.04,0.06,0.08,0.1,0.12},
                extra y tick labels={0.02,0.04,0.06,0.08,0.1,0.12},
                ytick style={color=black}]
        \addplot [thick, color0]
            table {%
                0 0.080300831438959
                1 0.0534964584449465
                2 0.04345744355692882
                3 0.0323173041072003
                4 0.0254528539297501
                5 0.0221728982678757
                6 0.0168120284100605
                7 0.0150541035595097
                8 0.0142533122920713
                9 0.0146548262465755
                10 0.0142687765111603
                11 0.0133859751626272
                12 0.0125473461323776
                13 0.0124531016447362
                14 0.0122719639874013
                15 0.011492895332179
                16 0.0116550302590742
                17 0.0117432497810486
                18 0.011257664250954
                19 0.0113862465008477
                20 0.0125142670321295
                21 0.0101831546616685
                22 0.0111995428925991
                23 0.00972847722434951
                24 0.009512524678857
                25 0.0094019869783397
                26 0.0091281277025362
                27 0.009674263513512
                28 0.00901587561302792
                29 0.00911587561302792
                30 0.00851587561302792
                31 0.00831587561302792
                32 0.00801587561302792
                33 0.00801587561302792
                34 0.00811587561302792
                35 0.0080587561302792
                36 0.00806887561302792
                37 0.0080987561302792
            
            };
            \addlegendentry{DNN\_scalar}
            \addplot [thick, color1]
            table {%
                0 0.1022226781068846
                1 0.0911920208877833
                2 0.0717080045670502
                3 0.0600201048703927
                4 0.0392319983218144
                5 0.0344504510484617
                6 0.028803313461154
                7 0.0274953422920648
                8 0.0269932203179644
                9 0.0279825467935905
                10 0.0266992175005357
                11 0.026977932848017
                12 0.0262059354285399
                13 0.0249797064177596
                14 0.0254347128213145
                15 0.0250388812391179
                16 0.0242895209889387
                17 0.0243792779914053
                18 0.0238461538049024
                19 0.0252858720432957
                20 0.0230487137669048
                21 0.0233629441501943
                22 0.024438886296495
                23 0.0229659146777997
                24 0.0239400564586546
                25 0.0227440737680717
                26 0.0248128873021258
                27 0.0224200455584498
                28 0.0223194206231671
                29 0.0211049949853303
                30 0.0218211473889399
                31 0.0216324643953403
                32 0.0214775482512221
                33 0.0216872822129449
                34 0.022434686038364
            };
            \addlegendentry{DCNN\_RISSP\_864x544}
            \addplot [thick, color2]
            table {%
                0 0.1179541133417332
                1 0.0931388400729929
                2 0.0742751642485614
                3 0.065023124020075
                4 0.0534602132709143
                5 0.052879985763044
                6 0.0478593352026564
                7 0.0461492881618575
                8 0.0385018953170887
                9 0.0383149240073662
                10 0.038422085264818
                11 0.0372351331161744
                12 0.0373118133509883
                13 0.037038516168716
                14 0.0345246074429547
                15 0.0349099459364356
                16 0.0337425232561214
                17 0.0345047118129733
                18 0.0339188641656277
                19 0.0331954163320613
                20 0.0328057004107061
                21 0.0336788410934059
                22 0.0345087828573792
                23 0.0321757554239774
                24 0.0331966048536837
                25 0.0302030192837697
                26 0.0314289875239596
                27 0.030959109939346
                28 0.0309217714169845
                29 0.0309849617306807
                30 0.0309074099271411
            };
            \addlegendentry{DCNN\_RISSP\_432x272}
            \addplot [thick, color3]
            table {%
                0 0.12879985763044
                1 0.1198593352026564
                2 0.101492881618575
                3 0.0901018953170887
                4 0.08249240073662
                5 0.076422085264818
                6 0.0730251331161744
                7 0.0671118133509883
                8 0.057038516168716
                9 0.0525246074429547
                10 0.04949099459364356
                11 0.0477425232561214
                12 0.0475047118129733
                13 0.0476188641656277
                14 0.0471954163320613
                15 0.0463388812391179
                16 0.0462895209889387
                17 0.0461792779914053
                18 0.0458461538049024
                19 0.0452858720432957
                20 0.0450487137669048
                21 0.0453629441501943
                22 0.045438886296495
                23 0.0453659146777997
                24 0.0452400564586546
                25 0.0442440737680717
                26 0.0448128873021258
                27 0.0444200455584498
                28 0.0443194206231671
                29 0.0443049949853303
                30 0.0448211473889399

            };
            \addlegendentry{DCNN\_RISSP\_216x136}
            \end{axis}
 \end{tikzpicture}
    \captionsetup{justification=centering}
    \caption{Training curve over validation set (Dataset~1)\label{fig:VAL1}. RISSP is short for rotor image and stator scalar parameters.}
    
\end{figure}
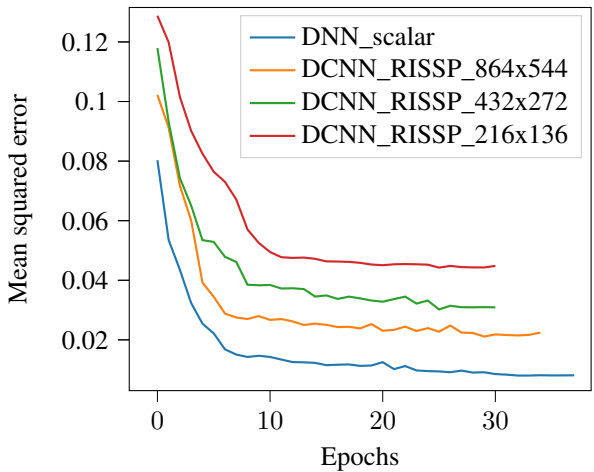
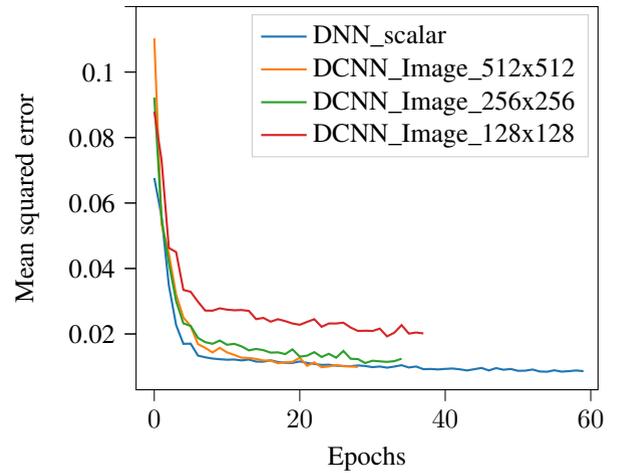
\begin{figure}
    \centering
\begin{tikzpicture}

\definecolor{color0}{rgb}{0.12156862745098,0.466666666666667,0.705882352941177}
\definecolor{color1}{rgb}{1,0.498039215686275,0.0549019607843137}
\definecolor{color2}{rgb}{0.172549019607843,0.627450980392157,0.172549019607843}
\definecolor{color3}{rgb}{0.83921568627451,0.152941176470588,0.156862745098039}

    \begin{axis}[legend cell align={left},
                width=0.9\linewidth,
                legend style={fill opacity=0.8, draw opacity=1, text opacity=1, draw=white!80!black},
                tick align=outside,
                tick pos=left,
                x grid style={white!69.0196078431373!black},
                xlabel={Epochs},
                xmin=-2.5, xmax=61,
                xtick style={color=black},
                y grid style={white!69.0196078431373!black},
                ylabel={Mean squared error},
                ymin=0.0029523, ymax=0.12,  
                yticklabels={},               
                extra y ticks={0.02,0.04,0.06,0.08,0.1},
                extra y tick labels={0.02,0.04,0.06,0.08,0.1},
                ytick style={color=black}]
    \addplot [thick, color0]
    table {%
    0 0.06764822842450259
    1 0.0558656161998517
    2 0.0348931660978677
    3 0.0227275269457382
    4 0.0169892291025829
    5 0.0170271095870308
    6 0.0133944482516103
    7 0.012862678087805
    8 0.0124894765419861
    9 0.0122854765841511
    10 0.0121312850956307
    11 0.0121880997145592
    12 0.0119150802902204
    13 0.0121802283470213
    14 0.011557389560904
    15 0.0115280500095752
    16 0.0119480544012954
    17 0.0114309260506154
    18 0.0111925940026917
    19 0.0111485751567153
    20 0.0116038609450101
    21 0.0111474814110024
    22 0.0108509563795202
    23 0.0105483263071617
    24 0.010593807175563
    25 0.0102990020321591
    26 0.010203224077378
    27 0.0101510075062937
    28 0.0103781413748976
    29 0.0102113167748086
    30 0.0098590108876427
    31 0.010060691731941
    32 0.00970485913073909
    33 0.0100379805381407
    34 0.0104512002473115
    35 0.00975521097503583
    36 0.0100601100283422
    37 0.00923591082102261
    38 0.00930556006132786
    39 0.00916289959045113
    40 0.00932438195098278
    41 0.00944073815187512
    42 0.00919593780154714
    43 0.00885520694243122
    44 0.00923550045430583
    45 0.00961575026380846
    46 0.00885174770235476
    47 0.00957186777115792
    48 0.00906631047055318
    49 0.00930230504079922
    50 0.00873685505683955
    51 0.00878195817887937
    52 0.00913487390665583
    53 0.00853148104444878
    54 0.00845940456092743
    55 0.00890270008847541
    56 0.00848998604655189
    57 0.00864501256563097
    58 0.0088427877125814
    59 0.00863952265729892
    };
    \addlegendentry{DNN\_scalar}
    \addplot [thick, color1]
    table {%
    0 0.110300831438959
    1 0.0534964584449465
    2 0.0445744355692882
    3 0.0320173041072003
    4 0.0250528539297501
    5 0.0223728982678757
    6 0.0169120284100605
    7 0.0157541035595097
    8 0.0143533122920713
    9 0.0157548262465755
    10 0.0143687765111603
    11 0.0135859751626272
    12 0.0127473461323776
    13 0.0126531016447362
    14 0.0123719639874013
    15 0.011892895332179
    16 0.0118550302590742
    17 0.0110432497810486
    18 0.011357664250954
    19 0.0114862465008477
    20 0.0127142670321295
    21 0.0102831546616685
    22 0.0112995428925991
    23 0.00982847722434951
    24 0.0100712524678857
    25 0.0105019869783397
    26 0.0101381277025362
    27 0.0100574263513512
    28 0.00992587561302792
    };
    \addlegendentry{DCNN\_Image\_512x512}
    \addplot [thick, color2]
    table {%
    0 0.0922226781068846
    1 0.0551920208877833
    2 0.0417080045670502
    3 0.0300201048703927
    4 0.0232319983218144
    5 0.0224504510484617
    6 0.018803313461154
    7 0.0174953422920648
    8 0.0169932203179644
    9 0.0179825467935905
    10 0.0166992175005357
    11 0.016977932848017
    12 0.0162059354285399
    13 0.0149797064177596
    14 0.0154347128213145
    15 0.0150388812391179
    16 0.0142895209889387
    17 0.0143792779914053
    18 0.0138461538049024
    19 0.0152858720432957
    20 0.0130487137669048
    21 0.0133629441501943
    22 0.014438886296495
    23 0.0129659146777997
    24 0.0139400564586546
    25 0.0127440737680717
    26 0.0148128873021258
    27 0.0124200455584498
    28 0.0123194206231671
    29 0.0111049949853303
    30 0.0118211473889399
    31 0.0116324643953403
    32 0.0114775482512221
    33 0.0116872822129449
    34 0.012434686038364
    };
    \addlegendentry{DCNN\_Image\_256x256}
    \addplot [thick, color3]
    table {%
    0 0.0879541133417332
    1 0.0731388400729929
    2 0.0462751642485614
    3 0.045023124020075
    4 0.0334602132709143
    5 0.032879985763044
    6 0.0298593352026564
    7 0.0271492881618575
    8 0.0271018953170887
    9 0.0278249240073662
    10 0.027422085264818
    11 0.0272351331161744
    12 0.0273118133509883
    13 0.027038516168716
    14 0.0245246074429547
    15 0.0249099459364356
    16 0.0237425232561214
    17 0.0245047118129733
    18 0.0239188641656277
    19 0.0231954163320613
    20 0.0228057004107061
    21 0.0236788410934059
    22 0.0245087828573792
    23 0.0221757554239774
    24 0.0231966048536837
    25 0.0232030192837697
    26 0.0234289875239596
    27 0.021959109939346
    28 0.0209217714169845
    29 0.0209849617306807
    30 0.0209074099271411
    31 0.0216304340113426
    32 0.0192581354772783
    33 0.0203785462839197
    34 0.0227084618244676
    35 0.0201086868852069
    36 0.0204374697155654
    37 0.0201652261262485
    };
    \addlegendentry{DCNN\_Image\_128x128}
    \end{axis}

\end{tikzpicture}
    \captionsetup{justification=centering}
    \caption{Training curve over validation set (Dataset~2)\label{fig:VAL2}}              
\end{figure}%

\begin{figure*}
	\centering      
    \input{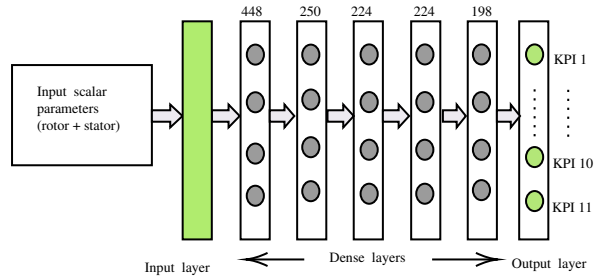}
    \caption{DNN: scalar parameter based meta-model}
    \label{fig:MLP}%
\end{figure*}

\begin{figure*}
    \centering
    \input{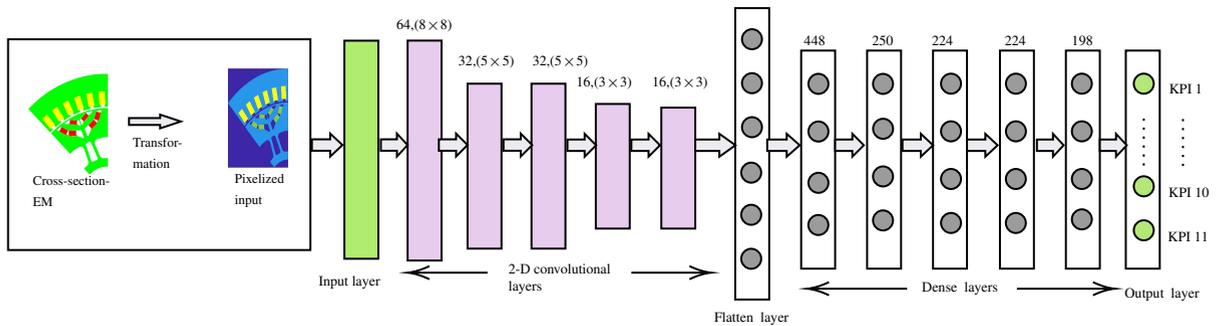}
    \captionsetup{justification=centering}
    \caption{DCNN: image based meta-model}
    \label{fig:DCNN_1}%
\end{figure*}

\begin{figure*}
    \centering
     \input{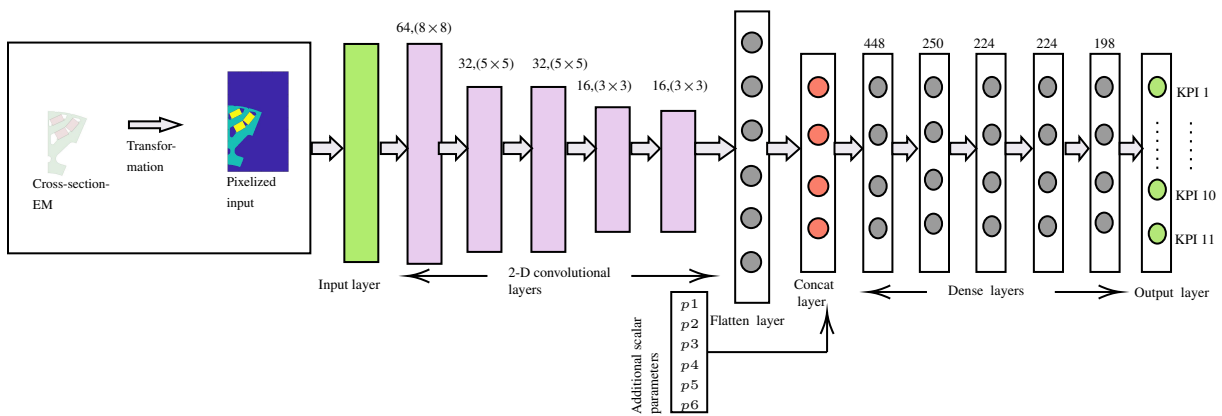}
	\captionsetup{justification=centering}
	\caption{DCNN with additional scalar input}
	\label{fig:DCNN_2}%
\end{figure*}

\begin{figure*}
    \centering
    \includegraphics[width=.90\linewidth]{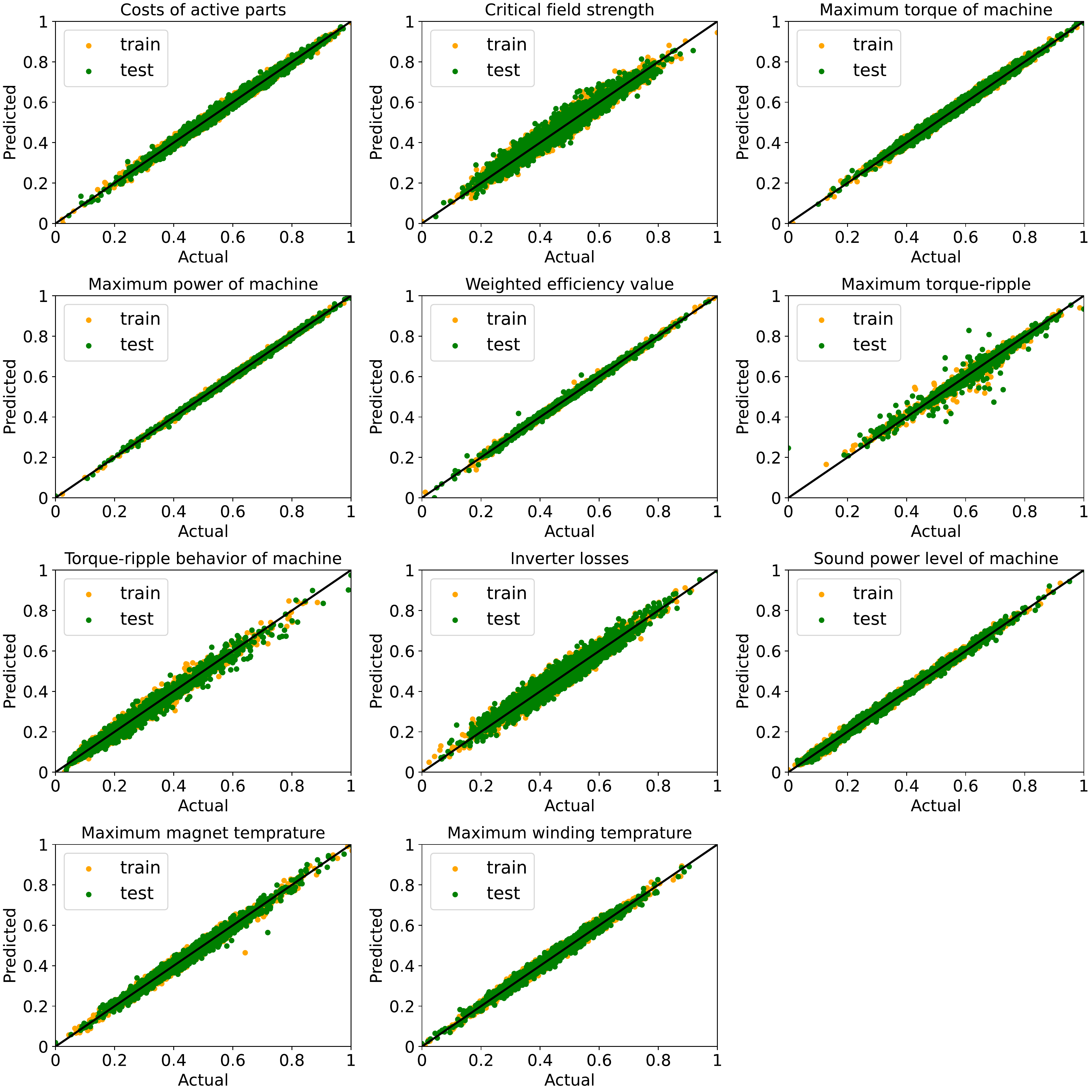}
    \vspace{-0.5em}
    \captionsetup{justification=centering}
    \caption{KPI prediction dataset~1 with scalar based meta-model}
    \label{fig:PPDS1}
\end{figure*}%
        
\begin{figure*}
	\centering
	\includegraphics[width=.90\linewidth]{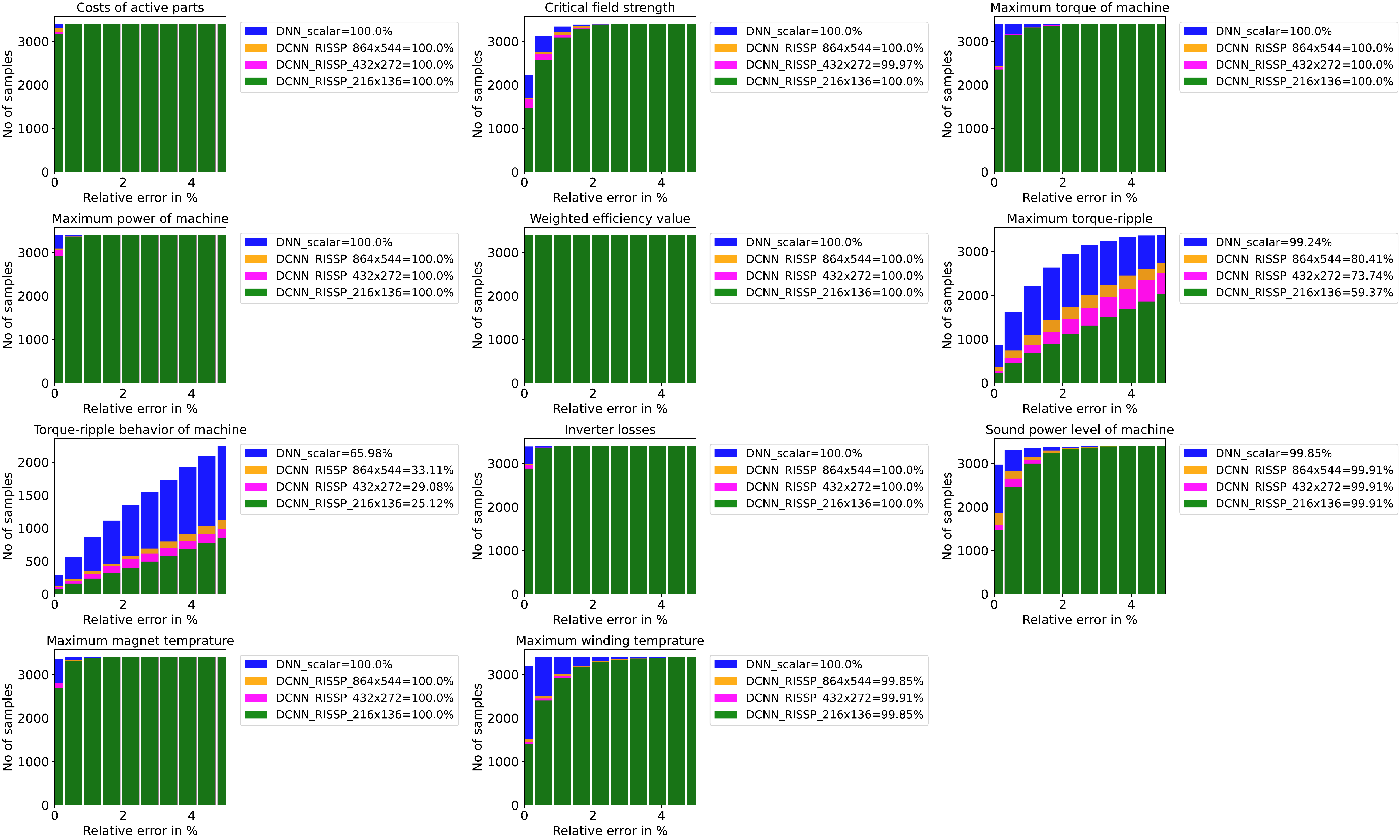}
	\vspace{-0.5em}
	\captionsetup{justification=centering}
	\caption{Cumulative accuracy plots of the KPI prediction with relative error $\varepsilon_{\textrm{mre}}<5\%$ dataset~1}
	\label{fig:REDS1}
\end{figure*}%
    
\begin{figure*}
	\centering
	\includegraphics[width=.90\linewidth]{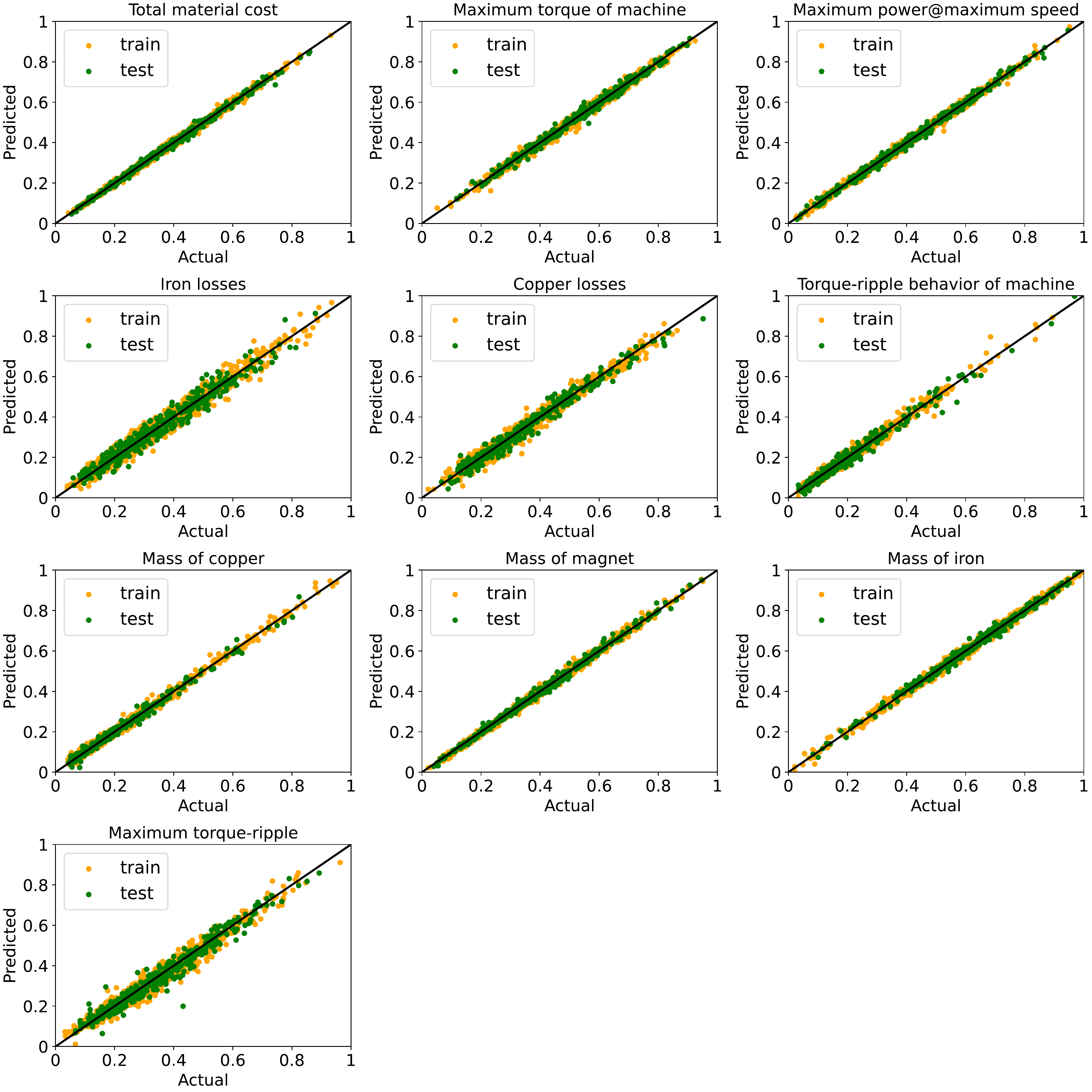}
	\vspace{-0.5em}
    \captionsetup{justification=centering}
    \caption{KPI prediction dataset~2 with scalar based meta-model}
    \label{fig:PPDS2}
\end{figure*}
        
\begin{figure*}
	\centering
    \includegraphics[width=.90\linewidth]{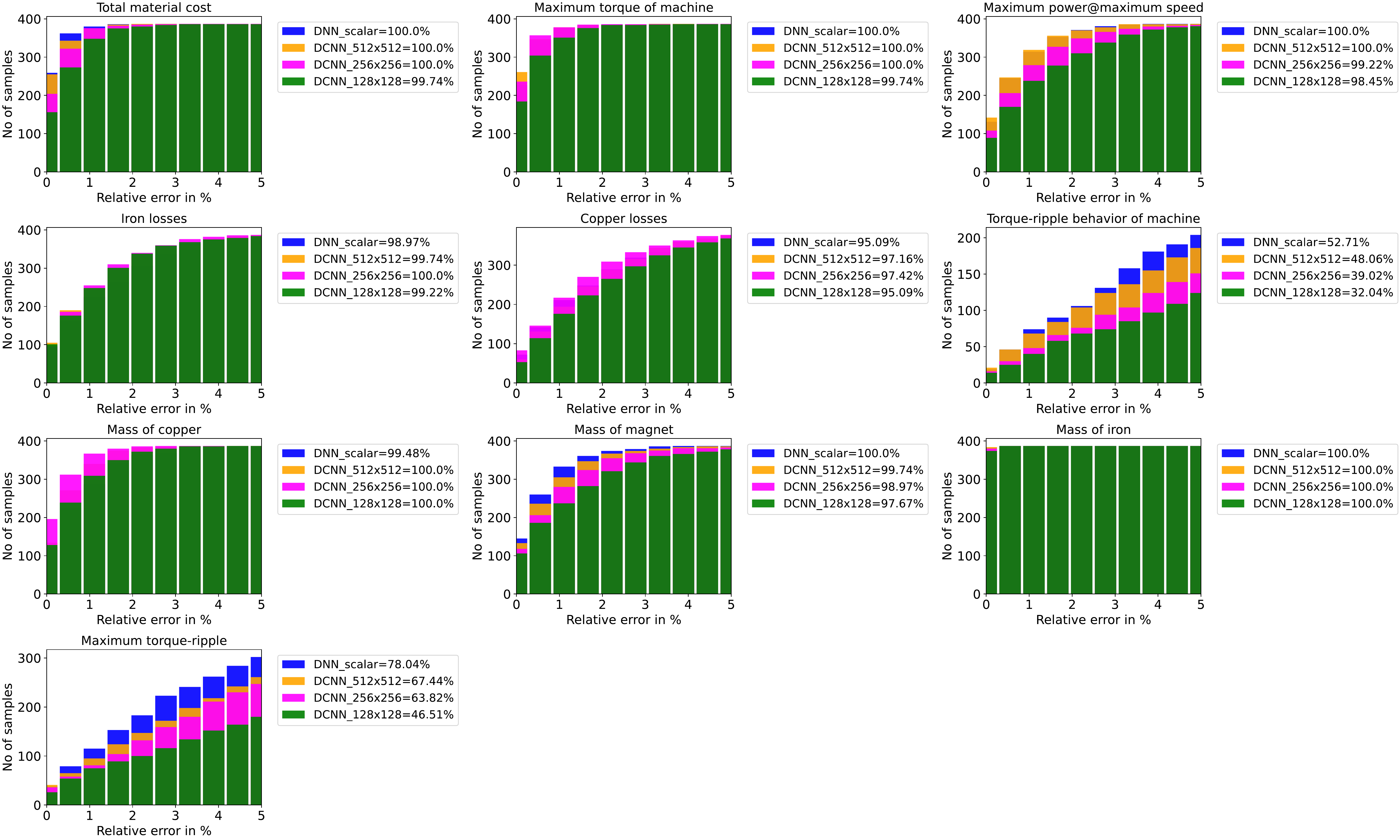}
    \vspace{-0.5em}
    \captionsetup{justification=centering}
	\caption{Cumulative accuracy plots of the KPI prediction with relative error $\varepsilon_{\textrm{mre}}<5\%$ dataset~2}
	\label{fig:REDS2}
\end{figure*}%

\begin{table}
    \centering
    \captionsetup{justification=centering}
    \caption{KPIs information dataset~2}
    \footnotesize
    \renewcommand{\arraystretch}{1.2}
    \begin{tabular}{@{}|r|l|c|@{}}
    \hline
               & \textbf{KPI}                      & \textbf{Unit} \\ \hline 
    $\kpi_{1}$ & Total cost                        & €             \\ \hline       
    $\kpi_{2}$ & Maximum torque of machine         & Nm            \\ \hline           
    $\kpi_{3}$ & Maximum power at maximum rpm      & KW            \\ \hline        
    $\kpi_{4}$ & Iron losses                       & W             \\ \hline              
    $\kpi_{5}$ & Copper losses                     & W             \\ \hline             
    $\kpi_{6}$ & Maximum torque ripple             & Nmp           \\ \hline     
    $\kpi_{7}$ & Mass of iron                     & Kg            \\ \hline     
    $\kpi_{8}$ & Mass of copper                    & Kg            \\ \hline      
    $\kpi_{9}$ & Mass of magnet                    & Kg            \\ \hline      
    $\kpi_{10}$ & Torque-ripple behavior of machine & -             \\ \hline
    \end{tabular}%
    \label{tab:KPIDS2}
\end{table}

\begin{table}
    \centering
    \captionsetup{justification=centering}
    \caption{Parameter detail dataset~2}
    \renewcommand{\arraystretch}{1.2}
    \footnotesize
    \begin{tabular}{@{}|l|l|r|r|c|@{}}
    \hline
    & \textbf{Parameter}                   & \textbf{Min.}       & \textbf{Max.}     & \textbf{Unit} \\ \hline
    $p_{ 1}$ & Angle of inner magnets      & $ 15\phantom{.0}$  & $ 40\phantom{.0}$  & degree        \\ \hline
    $p_{ 2}$ & Height of outer magnet      & $  3\phantom{.0}$    & $  7\phantom{.0}$& mm            \\ \hline
    $p_{ 3}$ & Pole angle of outer magnet  & $ 45\phantom{.0}$   & $ 80\phantom{.0}$ & degree        \\ \hline
    $ p_{4}$ & Height of tooth head        & $  4\phantom{.0}$   & $  7\phantom{.0}$ & mm            \\ \hline
    $ p_{5}$ & Rotor outer diameter        & $160\phantom{.0}$   & $170\phantom{.0}$ & mm            \\ \hline
    $p_{ 6}$ & Height of inner magnets     & $ 4\phantom{.0}$    & $  7\phantom{.0}$ & mm            \\ \hline
    $p_{ 7}$ & Width of inner magnets      & $  7.0$             & $ 11.5$           & mm            \\ \hline
    $p_{ 8}$ & Angle of inner magnets      & $ 28\phantom{.0}$   & $ 58\phantom{.0}$ & degree        \\ \hline
    $p_{ 9}$ & Width of outer magnet       & $  7\phantom{.0}$   & $ 12\phantom{.0}$ & mm            \\ \hline
    $p_{ 10}$& Angle of outer magnet       & $ 15\phantom{.0}$   & $ 38\phantom{.0}$ & degree        \\ \hline
    $p_{ 11}$ & Height of tooth head       & $ 12\phantom{.0}$   & $ 17\phantom{.0}$ & mm            \\ \hline
    $p_{12}$ & Tooth head width            & $  5\phantom{.0}$   & $  9\phantom{.0}$ & mm            \\ \hline
    \end{tabular}%
    \label{tab:PDS2}
\end{table}
%
\section{Network architecture and training}\label{sec:NAT}
The network architecture is being determined based on the form of input data used for the training. A deep neural network (DNN) or multilayer perceptron (MLP) architecture is derived for the meta-model based on scalar parameters. The DCNN is used for image-based data training. The network structure of \autoref{fig:DCNN_2} is also applied for the input combination of scalar and image-based data. The idea here is to explore how different forms of input data can affect predictive accuracy. The scalar parameters contain the full information of both rotor and the stator. While the combination of half-pole rotor cross-section image and stator parameters for dataset~1 sets the visual and scalar information of the electrical machine model. The images consist only of geometric details for the rotor and stator cross-sections for dataset~2. A trial and error approach was used to finalize three different candidates for each input. Details on the network architecture for each of these inputs are provided in the following subsections.

\subsection{DNN structure}\label{sec:mlp-model}

The MLP based DL model is shown in \autoref{fig:MLP} comprises of five dense layers. The network has an input layer with number of scalar parameters (rotor, stator) and output layer neurons with the number of target KPIs. The whole structure is defined as $56-448-250-224-224-198-11$ and $12-448-250-224-224-198-10$ for dataset~1 and dataset~2, respectively. The ELU activation function  has been chosen from the different non-linear activation functions between hidden layers \cite{clevert2016fast}.

\subsection{DCNN structure}\label{sec:dcnn-model}

The network architecture, as indicated in \autoref{fig:DCNN_1}, consists of two parts: convolution layers and dense layers. The objective of the convolutional layers is to extract spatially related features from the visual form of the input geometry. The dense layer section then uses this information to semantically project the discriminatory features that the convolution layers have extracted to predict KPIs in the final output layer. It must have sufficient capacity to successfully capture the complexity of the problem to train the network effectively. This functionality is accomplished by selecting an experience value by trial and error for the number of layers, the number of kernels($64, 32, 16$), the kernel size($8\times8$, $5\times5$, $3\times3$), and regularization parameters(learning rate).
As presented in \autoref{fig:DCNN_1}, there are five convolutional layers for down sampling. This network architecture is invariant to the input dimension. The dense layer structure after flattening the layer is the same as the model based on scalar parameters, see \autoref{fig:MLP}, to allow for reasonable comparisons.

\subsection{Multi-input DCNN structure}\label{sec:dcnn_model2}

The network architecture remains the same in this model as in the preceding \autoref{sec:dcnn-model}. The only difference being that the additional input layer is concatenated with the output of the convolution layer as shown in \autoref{fig:DCNN_2}. Detailed information is provided in \autoref{tab:STPDS1} for various scalar parameters of the stator geometry for dataset~1. The reason for this is to provide information on the missing stator geometry. This multi-input structure is only used for dataset~1 training as dataset~2 has already stator geometry information in image form.   

\subsection{Training process}\label{sec:training}

The data set is partitioned into training, validation and test sets. The networks are trained on the training set with back-propagation algorithm \cite{rumelhart1986learning} to learn any arbitrary mapping of input parameters to output KPIs. Hyper-parameters include the total number of training epochs maximum ($100$), batch size ($50$), early stopping over validation error(not decreasing continuously for $5$ epochs compared to the lowest error recorded so far during training), learning rate range ($0.001$-$0.0001$), metric evaluation (mean squared error), non-linear activation function (ELU), and optimizer (Adam) \cite{kingma2015adam}, network depth (count of hidden layers), and number of hidden units within each layer. Hyper-parameters also play a role in the model's performance to an extent, but it is difficult to fix specific values, so they are chosen randomly (trial and error) and kept constant for all the three model candidates to be trained. As mentioned in the section\ref{sec:model}, dataset~1 and dataset~2 consists of $n_1=68099$ and $n_2=7744$ samples, respectively. Approximately $90$\% of the total number of samples is used during the training process, while around {$5$}\% is reserved for validation and testing of both datasets, i.e.,for dataset 1: $n_{\textrm{train}}=61290$, $n_{\textrm{validation}}=3405$, $n_{\textrm{test}}=3404$ and for dataset~2 $n_{\textrm{train}}=6970$, $n_{\textrm{validation}}=387$, $n_{\textrm{test}}=387$. \autoref{fig:VAL1} and \autoref{fig:VAL2} shows training curve over the validation set for all the meta-models. The entire training process is carried out on a NVIDIA Quadro M4000 GPU. All the deep learning-based meta-models implemented using numerical computational library TensorFlow \cite{tensorflow2015-whitepaper}. Training on the scalar parameter based MLP model takes approximately $1$  minute to $4$ minutes for the dataset 2 and dataset 1 respectively. For the training of image based DCNN models, roughly $1$ to $8$ minutes per epoch (depending on image resolution, total number of training samples, and batch size), resulting in a total run time of $1$\,h to $6$\,h for both datasets. All meta-models take about $\sim1$\,ms/sample to evaluate new geometries that is much lower than the FE model, which calculates $6$\,h to $8$\,h on a single core CPU for one evaluation. If memory requirements are met, the advantage of parallelization over GPU during the training process can be further utilized with a higher number of geometries. The meta-model can then be trained in a short time, and thus the duration of training is memory-bound rather than compute-bound\cite{8661767}. Therefore, training time is greater for the large dataset(dataset 1).

\section{Results and analysis}\label{sec:res}

As the problem characterizes as a non-linear multi-output regression and all the KPIs are on different scales, the dimensionless mean relative error (MRE), \cite{guide1993guide}, i.e.,
 
\begin{equation} \label{eq:RMPE}
 \varepsilon_{\textrm{mre}}(\kpi_j)
 = 
 \frac{1}{n_{\textrm{test}}} 
 \sum_{i = 1}^{n_{\textrm{test}}}
 \frac{
    |
    \kpi_{j}^{(i)}
    -
    \tilde{\kpi}_{j}^{(i)}
    |
}{
    |
    \kpi_{j}^{(i)}
    |
}
\times 100
\end{equation}
is selected for the final evaluation of the $j$-th KPI on the $n_{\textrm{test}}$-dimensional validation data set. It quantifies how accurate the prediction $\tilde{\kpi}_j^{(i)}$ is compared to the true value $\kpi_j^{(i)}$ for a given parameter configuration $\mathbf{p}^{(i)}$. On the other hand the Pearson correlation coefficient (PCC)
\begin{equation}\label{eq:PCC}
  \varepsilon_\mathrm{pcc}(\kpi_j,\tilde{\kpi}_{j})
  =
  \frac{
  \sum_{i=1}^{n_{\textrm{test}}}
  (\kpi_j^{(i)}-\bar{\kpi_j}^{(i)})
  (\tilde{\kpi}_{j}^{(i)}-\bar{\tilde{\kpi}}_j^{(i)})
  }{%
    \sqrt{\sum_{i=1}^{n_{\textrm{test}}}
    (\kpi_j^{(i)}-\bar{\kpi_j}^{(i)})^2}
    \sqrt{\sum_{i=1}^{n_{\textrm{test}}}
    (\tilde{\kpi}_{j}^{(i)}-\bar{\tilde{\kpi}}_j^{(i)})^2}
  }
\end{equation}
 gives an idea of how the input parameters are mapped to the target output values \cite{ref1}. If the PCC is close to one between the predicted and the actual values, then the performance of the model is better.

\subsection{Evaluation of dataset~1}

    \autoref{fig:PPDS1} displays the predicted KPIs for the dataset~1 over their actual target values. An evaluation of all the KPIs is presented in \autoref{tab:ETDS1} over their mean values. A cumulative plot for below 5\% relative error is shown \autoref{fig:REDS1}. The MLP based model has input information on geometry parameters from both,the stator and the rotor, while the deep multi-input network\autoref{fig:DCNN_2} receives information in form of the half-pole rotor cross-section image and the stator geometry configuration. Training, test, validation set and hyper parameter settings remains constant during the training of all the meta-models. 
    
    The KPIs related to the torque behavior of the machine, e.g., $\kpi_{6}$, $\kpi_{7}$, have lower prediction performance compared to other KPIs with average $\!\varepsilon_\mathrm{mre}\!$, i.e.,  $\!\varepsilon_\mathrm{mre}\!$ $4.22\%$ and $1.28\%$, respectively, see \autoref{tab:ETDS1}.It can be observed that the average $\!\varepsilon_\mathrm{mre}\!$ over all the KPIs for the DNN is $0.64\%$ which is much lower than the best performing multi-input DCNN model (with resolution $544 \times 864$\,pixel) with average $\!\varepsilon_\mathrm{mre}\!$ of $1.43\%$. The DCNN based multi-input meta-model is trained and evaluated with three different image resolutions. The network architecture is shown in \autoref{fig:DCNN_2}. The network with the higher resolution image data($864\times544$) has average $\!\varepsilon_\mathrm{mre}\!$ $1.43\%$ over all the KPIs that is $\sim 12.05\% $ and $\sim 24.96\% $ lower than the input data with image resolution $432\times272$ and $216\times136$, respectively. Detailed result is described in \autoref{tab:ETDS1}. However, as a consequence of the higher resolution, the network takes twice or sometimes even more time to train as compared the lower one, i.e., approx. $1$h ($136\times216$\,pixel) vs. approx. $2$h ($272\times472$\,pixel) with average over all KPIs $\!\varepsilon_\mathrm{mre}\!$ lower $\sim 14.68\%$ for high resolution, so it is trade-off between the training time of meta-model and performance.
    
    \begin{table}
            \centering
            \captionsetup{justification=centering}
            \caption{Evaluation summary dataset~1}
            \renewcommand{\arraystretch}{1.2}
            \resizebox{\columnwidth}{!}{%
            \footnotesize
            \begin{tabular}{@{}|l|r|r|r|r|r|r|r|r|@{}}
            \hline
            \multirow{2}{*}{~} &
              \multicolumn{2}{p{1.6cm}|}{\centering\textbf{DNN}} &
              \multicolumn{2}{p{1.6cm}|}{\centering\textbf{DCNN}\newline$544\times864$} &
              \multicolumn{2}{p{1.6cm}|}{\centering\textbf{DCNN}\newline$272\times432$} &
              \multicolumn{2}{p{1.6cm}|}{\centering\textbf{DCNN}\newline$136\times216$} \\ 
             &
              $\!\varepsilon_\mathrm{mre}\!$ &
              $\!\varepsilon_\mathrm{pcc}\!$ &
              $\!\varepsilon_\mathrm{mre}\!$ &
              $\!\varepsilon_\mathrm{pcc}\!$ &
              $\!\varepsilon_\mathrm{mre}\!$ &
              $\!\varepsilon_\mathrm{pcc}\!$ &
              $\!\varepsilon_\mathrm{mre}\!$ &
              $\!\varepsilon_\mathrm{pcc}\!$ \\ 
             \hline
            $\kpi_{ 1}$ & $0.12$ & $0.99$ & $0.17$ & $0.99$ & $ 0.20$ & $0.99$ & $ 0.22$ & $0.98$\\ \hline          
            $\kpi_{ 2}$ & $0.44$ & $0.97$ & $0.60$ & $0.96$ & $ 0.64$ & $0.97$ & $ 0.70$ & $0.95$\\ \hline      
            $\kpi_{ 3}$ & $0.12$ & $0.99$ & $0.41$ & $0.98$ & $ 0.42$ & $0.93$ & $ 0.42$ & $0.93$\\ \hline      
            $\kpi_{ 4}$ & $0.05$ & $0.98$ & $0.24$ & $0.98$ & $ 0.25$ & $0.95$ & $ 0.27$ & $0.94$\\ \hline     
            $\kpi_{ 5}$ & $0.01$ & $0.94$ & $0.05$ & $0.92$ & $ 0.05$ & $0.90$ & $ 0.06$ & $0.89$\\ \hline  
            $\kpi_{ 6}$ & $1.28$ & $0.98$ & $2.99$ & $0.97$ & $ 3.55$ & $0.96$ & $ 4.77$ & $0.95$\\ \hline
            $\kpi_{ 7}$ & $4.22$ & $0.95$ & $9.4$ & $0.94$ & $ 10.69$ & $0.92$ & $ 12.34$& $0.89$\\ \hline   
            $\kpi_{ 8}$ & $0.13$ & $0.98$ & $0.26$ & $0.98$ & $ 0.26$ & $0.98$ & $ 0.28$ & $0.96$\\ \hline 
            $\kpi_{ 9}$ & $0.29$ & $0.96$ & $0.55$ & $0.95$ & $ 0.71$ & $0.94$ & $ 0.76$ & $0.94$\\ \hline 
            $\kpi_{10}$ & $0.16$ & $0.98$ & $0.32$ & $0.94$ & $ 0.35$ & $0.93$ & $ 0.35$ & $0.91$\\ \hline          
            $\kpi_{11}$ & $0.21$ & $0.96$ & $0.76$ & $0.95$ & $ 0.79$ & $0.95$ & $ 0.82$ & $0.95$\\ \hline                   
            \end{tabular}%
            }
            \label{tab:ETDS1}
    \end{table}

\subsection{Evaluation of dataset~2}

    Dataset~2 has fewer samples which are more uniformly distributed than dataset~1 and has a different set of KPIs as well. Prediction plot and evaluations are exhibited in \autoref{fig:PPDS2} and \autoref{tab:ETDS2}, respectively. A cumulative plot for the error is illustrated in \autoref{fig:REDS2}. The KPIs related to the machine torque ($\kpi_{6}$ and $\kpi_{10}$) have the down predicatibility than other KPIs with average $\!\varepsilon_\mathrm{mre}\! $ $5.9\%$ and $3.47\%$ over all the KPIs. It is also evident from the results that the prediction accuracy improves with image accuracy for this dataset, too. The scalar parameter based meta-model has average $\!\varepsilon_\mathrm{mre}\!$ $1.66\%$ which is $7.45\%$, $15.78\%$, $34.28\%$ lower than the image based DCNN $512\times512$, image based DCNN $256\times256$, and $128\times128$, respectively. Here, it is important to note that dataset~2 has a less scalar parameters, i.e., $12$ and cross-section of EM consists geometry information of one full pole rotor and stator. 
    
\subsection{DNN and Gaussian process regression for parameter based meta-models}
\begin{figure}
        \centering
        \begin{tikzpicture}

\definecolor{color0}{rgb}{0.12156862745098,0.466666666666667,0.705882352941177}
\definecolor{color1}{rgb}{1,0.498039215686275,0.0549019607843137}

\begin{axis}[
legend cell align={left},
legend style={fill opacity=0.8, draw opacity=1, text opacity=1, draw=white!80!black},
tick align=outside,
tick pos=left,
x grid style={white!69.0196078431373!black},
xlabel={KPIs},
xmin=-0.78, xmax=9.78,
xtick style={color=black},
xtick={0,1,2,3,4,5,6,7,8,9},
xticklabels={\(\displaystyle y_{1}\),\(\displaystyle y_{2}\),\(\displaystyle y_{3}\),\(\displaystyle y_{4}\),\(\displaystyle y_{5}\),\(\displaystyle y_{6}\),\(\displaystyle y_{7}\),\(\displaystyle y_{8}\),\(\displaystyle y_{9}\),\(\displaystyle y_{10}\)},
y grid style={white!69.0196078431373!black},
ylabel={Mean relative error (\%)},
ymin=0, ymax=14.406,
ytick style={color=black}
]
\draw[draw=none,fill=color0] (axis cs:-0.3,0) rectangle (axis cs:0,1.37);
\addlegendimage{area legend,fill=color0};
\addlegendentry{GPR}

\draw[draw=none,fill=color0] (axis cs:0.7,0) rectangle (axis cs:1,1.69);
\draw[draw=none,fill=color0] (axis cs:1.7,0) rectangle (axis cs:2,2.54);
\draw[draw=none,fill=color0] (axis cs:2.7,0) rectangle (axis cs:3,2.88);
\draw[draw=none,fill=color0] (axis cs:3.7,0) rectangle (axis cs:4,3.71);
\draw[draw=none,fill=color0] (axis cs:4.7,0) rectangle (axis cs:5,13.72);
\draw[draw=none,fill=color0] (axis cs:5.7,0) rectangle (axis cs:6,3.22);
\draw[draw=none,fill=color0] (axis cs:6.7,0) rectangle (axis cs:7,2.64);
\draw[draw=none,fill=color0] (axis cs:7.7,0) rectangle (axis cs:8,0.38);
\draw[draw=none,fill=color0] (axis cs:8.7,0) rectangle (axis cs:9,8.61);
\draw[draw=none,fill=color1] (axis cs:2.77555756156289e-17,0) rectangle (axis cs:0.3,0.42);
\addlegendimage{area legend,fill=color1};
\addlegendentry{DNN}

\draw[draw=none,fill=color1] (axis cs:1,0) rectangle (axis cs:1.3,0.51);
\draw[draw=none,fill=color1] (axis cs:2,0) rectangle (axis cs:2.3,0.91);
\draw[draw=none,fill=color1] (axis cs:3,0) rectangle (axis cs:3.3,1.52);
\draw[draw=none,fill=color1] (axis cs:4,0) rectangle (axis cs:4.3,1.83);
\draw[draw=none,fill=color1] (axis cs:5,0) rectangle (axis cs:5.3,5.9);
\draw[draw=none,fill=color1] (axis cs:6,0) rectangle (axis cs:6.3,1.06);
\draw[draw=none,fill=color1] (axis cs:7,0) rectangle (axis cs:7.3,0.84);
\draw[draw=none,fill=color1] (axis cs:8,0) rectangle (axis cs:8.3,0.13);
\draw[draw=none,fill=color1] (axis cs:9,0) rectangle (axis cs:9.3,3.47);
\draw (axis cs:-0.05,1.37) ++(0pt,1pt) node[
  scale=0.5,
  anchor=south,
  text=black,
  rotate=0.0
]{1.37};
\draw (axis cs:0.95,1.69) ++(0pt,1pt) node[
  scale=0.5,
  anchor=south,
  text=black,
  rotate=0.0
]{1.69};
\draw (axis cs:1.95,2.54) ++(0pt,1pt) node[
  scale=0.5,
  anchor=south,
  text=black,
  rotate=0.0
]{2.54};
\draw (axis cs:2.95,2.88) ++(0pt,1pt) node[
  scale=0.5,
  anchor=south,
  text=black,
  rotate=0.0
]{2.88};
\draw (axis cs:3.95,3.71) ++(0pt,1pt) node[
  scale=0.5,
  anchor=south,
  text=black,
  rotate=0.0
]{3.71};
\draw (axis cs:4.95,13.72) ++(0pt,1pt) node[
  scale=0.5,
  anchor=south,
  text=black,
  rotate=0.0
]{13.72};
\draw (axis cs:5.95,3.22) ++(0pt,1pt) node[
  scale=0.5,
  anchor=south,
  text=black,
  rotate=0.0
]{3.22};
\draw (axis cs:6.95,2.64) ++(0pt,1pt) node[
  scale=0.5,
  anchor=south,
  text=black,
  rotate=0.0
]{2.64};
\draw (axis cs:7.95,0.38) ++(0pt,1pt) node[
  scale=0.5,
  anchor=south,
  text=black,
  rotate=0.0
]{0.38};
\draw (axis cs:8.95,8.61) ++(0pt,1pt) node[
  scale=0.5,
  anchor=south,
  text=black,
  rotate=0.0
]{8.61};
\draw (axis cs:0.25,0.42) ++(0pt,1pt) node[
  scale=0.5,
  anchor=south,
  text=black,
  rotate=0.0
]{0.42};
\draw (axis cs:1.25,0.51) ++(0pt,1pt) node[
  scale=0.5,
  anchor=south,
  text=black,
  rotate=0.0
]{0.51};
\draw (axis cs:2.25,0.91) ++(0pt,1pt) node[
  scale=0.5,
  anchor=south,
  text=black,
  rotate=0.0
]{0.91};
\draw (axis cs:3.25,1.52) ++(0pt,1pt) node[
  scale=0.5,
  anchor=south,
  text=black,
  rotate=0.0
]{1.52};
\draw (axis cs:4.25,1.83) ++(0pt,1pt) node[
  scale=0.5,
  anchor=south,
  text=black,
  rotate=0.0
]{1.83};
\draw (axis cs:5.25,5.9) ++(0pt,1pt) node[
  scale=0.5,
  anchor=south,
  text=black,
  rotate=0.0
]{5.9};
\draw (axis cs:6.25,1.06) ++(0pt,1pt) node[
  scale=0.5,
  anchor=south,
  text=black,
  rotate=0.0
]{1.06};
\draw (axis cs:7.25,0.84) ++(0pt,1pt) node[
  scale=0.5,
  anchor=south,
  text=black,
  rotate=0.0
]{0.84};
\draw (axis cs:8.25,0.13) ++(0pt,1pt) node[
  scale=0.5,
  anchor=south,
  text=black,
  rotate=0.0
]{0.13};
\draw (axis cs:9.25,3.47) ++(0pt,1pt) node[
  scale=0.5,
  anchor=south,
  text=black,
  rotate=0.0
]{3.47};
\end{axis}

\end{tikzpicture}
        \captionsetup{justification=centering}
        \caption{Dataset 2 : KPIs performance comparision for parameter based meta-model  \label{fig:GPR_DNN}}              
    \end{figure}
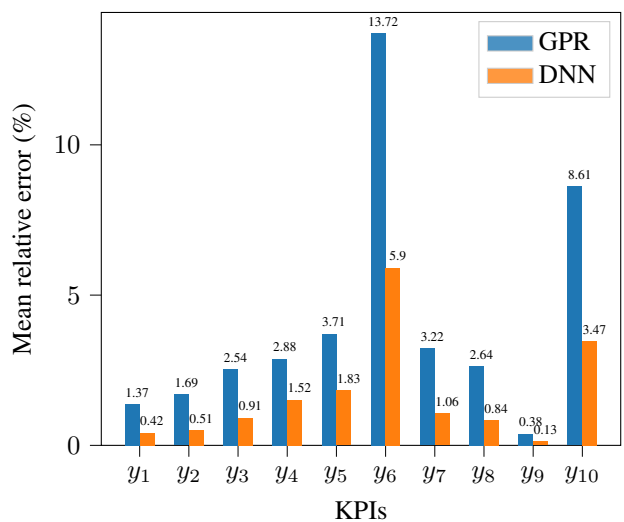

This work focuses on meta-modeling approaches based on deep learning as they promise to efficiently treat image based data \cite{krizhevsky2012imagenet}. However, to allow for a comparison with other state of the art approaches, also Kriging, or more precisely, Gaussian process regression (GPR), see e.g. \cite{10.5555/1162254}), is applied in the case of parameter based learning. The GPR meta-model is trained using the sci-kit-learn library \cite{JMLR:v12:pedregosa11a,sklearn_api} with its default settings, i.e., radial-basis function kernel  (RBF 1.0) and default optimizer L-BGFS-B \cite{doi:10.1137/0916069,10.1145/2049662.2049669}. The RBF kernel is often used in practice, however, a more rigorous quantitative comparison should take other kernels into account but this is beyond the scope of this paper. For dataset 1, all the KPIs have similar performance except torque related KPIs for which the DNN-based meta-model has around $50\%$ less error than the GPR based meta-model. \autoref{fig:GPR_DNN} illustrates the outcomes of both approaches concerning dataset 2. The results show that the DNN-based meta-model for all the KPIs clearly outperforms the GPR-based meta-model. The training time for the GPR based meta-model is approx. $10$ times larger than for the DNN. Dataset 1 has a large number of samples and higher dimensional input space compared to dataset 2, therefore the GPR-based meta-model was trained in two separate runs due to memory constraints.
      
    \begin{table}
        \captionsetup{justification=centering}
        \caption{Evaluation summary dataset~2}
        \renewcommand{\arraystretch}{1.2}
        \resizebox{\columnwidth}{!}{%
        \footnotesize
        \begin{tabular}{@{}|l|r|r|r|r|r|r|r|r|@{}}
        \hline
        \multirow{2}{*}{~} &
        \multicolumn{2}{p{1.6cm}|}{\centering\textbf{DNN}} &
        \multicolumn{2}{p{1.6cm}|}{\centering\textbf{DCNN}\newline$512\times512$} &
        \multicolumn{2}{p{1.6cm}|}{\centering\textbf{DCNN}\newline$256\times256$} &
        \multicolumn{2}{p{1.6cm}|}{\centering\textbf{DCNN}\newline$128\times128$} \\ 
        &
        $\!\varepsilon_\mathrm{mre}\!$ &
        $\!\varepsilon_\mathrm{pcc}\!$ &
        $\!\varepsilon_\mathrm{mre}\!$ &
        $\!\varepsilon_\mathrm{pcc}\!$ &
        $\!\varepsilon_\mathrm{mre}\!$ &
        $\!\varepsilon_\mathrm{pcc}\!$ &
        $\!\varepsilon_\mathrm{mre}\!$ &
        $\!\varepsilon_\mathrm{pcc}\!$ \\ 
        \hline
        $\kpi_{ 1}$ & $0.42$ & $1.00$ & $0.47$ & $1.00$ & $0.56$ & $1.00$ & $0.78$ & $0.99$ \\ \hline    
        $\kpi_{ 2}$ & $0.51$ & $1.00$ & $0.45$ & $1.00$ & $0.46$ & $0.99$ & $0.68$ & $0.98$ \\ \hline       
        $\kpi_{ 3}$ & $0.91$ & $1.00$ & $0.90$ & $1.00$ & $1.16$ & $0.99$ & $1.49$ & $0.99$ \\ \hline     
        $\kpi_{ 4}$ & $1.52$ & $0.98$ & $1.33$ & $0.99$ & $1.26$ & $0.99$ & $1.31$ & $0.96$ \\ \hline    
        $\kpi_{ 5}$ & $1.83$ & $0.99$ & $1.82$ & $0.99$ & $1.64$ & $0.99$ & $2.06$ & $0.97$ \\ \hline    
        $\kpi_{ 6}$ & $5.9$ & $0.98$ & $6.70$ & $0.98$ &  $7.99$ & $0.96$ & $9.67$ & $0.97$ \\ \hline 
        $\kpi_{ 7}$ & $1.06$ & $0.99$ & $0.78$ & $0.99$ & $0.62$ & $0.98$ & $0.93$ & $0.94$ \\ \hline   
        $\kpi_{ 8}$ & $0.84$ & $1.00$ & $0.97$ & $1.00$ & $1.15$ & $1.00$ & $1.53$ & $0.98$ \\ \hline        
        $\kpi_{ 9}$ & $0.13$ & $1.00$ & $0.14$ & $1.00$ & $0.14$ & $1.00$ & $0.17$ & $0.99$ \\ \hline       
        $\kpi_{10}$ & $3.47$ & $0.98$ & $4.36$ & $0.98$ & $4.72$ & $0.97$ & $6.58$ & $0.95$ \\ \hline      
        \end{tabular}%
        }
        \label{tab:ETDS2}
    \end{table}
    
\section{Conclusion}\label{sec:conclusion}
This contribution shows that the deep learning meta-models can be effectively used to approximate a large number cross-domain KPIs in a high dimensional parameter space. The data for demonstration is taken from a real-world industrial design workflow. The meta-model enables us to predict KPIs for new geometries at much lower computational costs. The prediction performance depends on how accurately the input information can be mapped to the target KPIs. The mapping accuracy relies mainly on two factors, hyper-parameter settings and the precision of the input data. In this work, hyper-parameter settings are the same for all meta-models and the focus is on the precision of the input data. This paper proposes two models for parameter and image-based learning. The image-based approach increases the flexibility and re-usability of the model for example in the case of a reparametrization. Our results show that it performs close to scalar-parameter-based models if the pixel resolution of the training data is sufficient. Future work will make use of the meta-model it many-query-scenarios, e.g., uncertainty quantification or multi-objective optimization.

\printbibliography
\EOD
\end{document}